\documentclass{article}

\usepackage{microtype}
\usepackage{graphicx}
\usepackage{booktabs} 

\usepackage{hyperref}

\usepackage[accepted]{icml2025}

\usepackage{amsmath}
\usepackage{amssymb}
\usepackage{mathtools}
\usepackage{amsthm}

\usepackage[capitalize,noabbrev]{cleveref}

\theoremstyle{plain}

\theoremstyle{definition}

\theoremstyle{remark}

\usepackage[textsize=tiny]{todonotes}

\usepackage{url}
\usepackage{multirow}
\usepackage{booktabs} 
\usepackage{array}
\usepackage{wrapfig}
\usepackage{colortbl}
\usepackage{tabularx}
\usepackage{subcaption}

\icmltitlerunning{CogMath: Assessing LLMs' Authentic Mathematical Ability from a Human Cognitive Perspective}

\begin{document}

\twocolumn[
\icmltitle{CogMath: Assessing LLMs' Authentic Mathematical Ability from \\ a Human Cognitive Perspective}



\icmlsetsymbol{equal}{*}

\begin{icmlauthorlist}
\icmlauthor{Jiayu Liu}{1,2}
\icmlauthor{Zhenya Huang}{2,3}
\icmlauthor{Wei Dai}{1}
\icmlauthor{Cheng Cheng}{2,3}
\icmlauthor{Jinze Wu}{2,4}
\icmlauthor{Jing Sha}{2,4}
\icmlauthor{Song Li}{2,4}
\icmlauthor{Qi Liu}{1,2}
\icmlauthor{Shijin Wang}{2,4}
\icmlauthor{Enhong Chen}{2,3}
\end{icmlauthorlist}

\icmlaffiliation{1}{School of Artificial Intelligence and Data Science, University of Science and Technology of China}
\icmlaffiliation{2}{State Key Laboratory of Cognitive Intelligence}
\icmlaffiliation{3}{School of Computer Science and Technology, University of Science and Technology of China}
\icmlaffiliation{4}{iFLYTEK AI Research}

\icmlcorrespondingauthor{Enhong Chen}{cheneh@ustc.edu.cn}

\icmlkeywords{Machine Learning, ICML}

\vskip 0.3in
]



\printAffiliationsAndNotice{} 

\begin{abstract}
Although large language models (LLMs) show promise in solving complex mathematical tasks, existing evaluation paradigms rely solely on a coarse measure of overall answer accuracy, which are insufficient for assessing their authentic capabilities. In this paper, we propose \textbf{CogMath}, which comprehensively assesses LLMs' mathematical abilities through the lens of human cognition. Specifically, inspired by psychological theories, CogMath formalizes human reasoning process into 3 stages: \emph{problem comprehension}, \emph{problem solving}, and \emph{solution summarization}. Within these stages, we investigate perspectives such as numerical calculation, knowledge, and counterfactuals, and design a total of 9 fine-grained evaluation dimensions. In each dimension, we develop an ``\emph{Inquiry}-\emph{Judge}-\emph{Reference}'' multi-agent system to generate inquiries that assess LLMs' mastery from this dimension. An LLM is considered to truly master a problem only when excelling in all inquiries from the 9 dimensions. By applying CogMath on three benchmarks, we reveal that the mathematical capabilities of 7 mainstream LLMs are overestimated by 30\%-40\%. Moreover, we locate their strengths and weaknesses across specific stages/dimensions, offering in-depth insights to further enhance their reasoning abilities.
\end{abstract}

\section{Introduction}
The rise of large language models (LLMs) has marked a pivotal moment in artificial intelligence. Particularly within the realm of mathematical reasoning, these models have made breakthroughs in solving complex mathematical problems~\cite{wei2022chain,xue2024decompose}. For example, GPT-4 has achieved over 75\% accuracy on the high school competition-level MATH dataset~\citep{hendrycks2021measuring}. More recently, the OpenAI-o1 model has surpassed 70\% accuracy on the AIME math competition, placing it at a level comparable to the top 500 US high school students~\footnote{https://openai.com/index/openai-o1-mini-advancing-cost-efficient-reasoning/}. This remarkable progress has not only redefined the potential of AI in mathematics but also spurred a growing body of research dedicated to evaluating and understanding the mathematical proficiency of these models.

To assess the mathematical ability of LLMs, numerous benchmarks have been proposed~\cite{li2024perteval}. For instance, E-GSM~\citep{xu2024can} introduces problems across four length ranges to assess LLMs' generalization with respect to input length. GSM-Plus~\citep{li2024gsm} introduces eight variants of GSM8K dataset~\citep{cobbe2021training} to investigate the robustness of LLMs. MPA~\citep{zhu2024dynamic} rewrites existing datasets based on five principles to address the effects of data contamination. However, on one hand, some benchmarks are overly task-specific, focusing on narrow aspects of reasoning (e.g., long-text understanding) by designing specialized problem types (e.g., long-text questions). On the other hand, they rely on a coarse accuracy metric that overlooks the detailed reasoning processes. As a result, they are unable to fully grasp the entire spectrum of mathematical capabilities that LLMs possess.

In this paper, we propose \textbf{CogMath}, which offers a scientific and comprehensive evaluation of LLMs' mathematical abilities by delving into the cognitive stages of human reasoning processes. Specifically, psychological research points out that human mathematical reasoning typically involves three stages~\citep{schoenfeld2014mathematical,lesh2003foundations,dehaene1999sources}: \emph{problem comprehension}, \emph{problem solving}, and \emph{solution summarization}. Aligned with these stages, we design nine evaluation dimensions covering key aspects such as computation, knowledge, and counterfactual reasoning. For example, in \emph{problem comprehension} stage, we assess the model's ability to handle different formulations of the same problem. In \emph{problem solving} stage, we break down the solution into three orthogonal components: problem-solving strategy, numerical computation, and knowledge application, and evaluate LLMs in each aspect independently. In \emph{solution summarization} stage, we go beyond traditional forward evaluation by introducing intermediate-step questions and backward reasoning tasks, testing whether the model can trace back through its reasoning pathway. 

To carry out the evaluation in each dimension, we also design an ``\emph{Inquiry-Judge-Reference}'' multi-agent system: the \emph{Inquiry} agent poses a dimension-specific inquiry about the problem, the \emph{Judge} agent refines the inquiry to ensure its quality, and the \emph{Reference} agent provides a correct answer as a standard to assess the LLM's performance. Unlike traditional evaluation paradigms that rely solely on an answer accuracy, CogMath considers an LLM to truly master a problem only after excelling in all inquiries in 9 dimensions. 

We apply CogMath on the most representative mathematical benchmarks GSM8K and MATH, along with an additional dataset we collected, MExam, which is composed of real exam tests that cover the full K-12 curriculum. Then, we evaluate 7 mainstream LLMs including GPT-4~\citep{achiam2023gpt}, GPT-3.5-Turbo~\citep{chatgpt2023}, Gemini-1.5-Flash~\citep{team2023gemini}, DeepSeek-V2.5~\cite{liu2024deepseek}, Llama3-8B~\citep{meta2024introducing}, Llama2-13B~\citep{touvron2023llama}, and Mixtral-8x7B-Instruct~\citep{mistral2023mixtral}. Our key experimental findings are as follows~\footnote{Our code and data are available at \url{https://github.com/Ljyustc/CogMath}.}:
\begin{itemize}
  \item The authentic mathematical capabilities of current LLMs are overestimated by 30\%-40\%. For instance, GPT-4 has truly mastered only 39.7\% and 67.1\% of the problems in MATH and GSM8K datasets, respectively. Moreover, this overestimation is not solely attributable to data contamination, but rather to an excessive imitation of superficial patterns of reasoning.
  \item We locate the deficiency stage of LLMs. Weaker models (e.g., Llama2-13B) still struggle in \emph{problem comprehension} stage, while stronger models (e.g., DeepSeek-V2.5) face challenges primarily in \emph{problem solving} stage, particularly in their mastery of knowledge.
  \item Confronted with a counterfactual setting, LLMs may exhibit an inherent ``over-correction'' behavior, automatically aligning with patterns from the training data.
  \item Existing prompting techniques, such as CoT and ICL, fail to truly improve the mathematical reasoning capabilities of LLMs.
\end{itemize}
\section{Related Work}
\subsection{Large Language Models}
Large language models (LLMs) have significantly advanced the field of natural language processing (NLP)~\cite{min2023recent,zhao2024explainability,liu2025knowledge,zhao2024comprehensive}. Models like OpenAI-o1, GPT-4~\citep{achiam2023gpt}, and GPT-3.5-Turbo~\citep{chatgpt2023} have set new performance milestones across numerous NLP tasks, such as sentiment classification~\citep{zhang2024sentiment}, question answering~\citep{hendrycks2021measuring}, and translation~\citep{wang2023document}. To further enhance their reasoning and problem-solving abilities, several advanced techniques have been introduced. Among them, Chain-of-Thought (CoT)~\citep{wei2022chain}, Tree-of-Thought (ToT)~\citep{yao2024tree}, and Graph-of-Thought (GoT)~\citep{besta2024graph} simulate structured and logical reasoning paths using chains, trees, and graphs, respectively, allowing models to handle multi-step problems more effectively. In-Context Learning (ICL)~\citep{dong2022survey} enables a model to learn from a few examples to generalize and solve unseen problems. In addition, there are many other key techniques, such as self-consistency~\citep{wangself}, Retrieval-Augmented Generation (RAG)~\citep{chen2024benchmarking}, and tool learning~\citep{ma2025automated}. We refer the readers to a more detailed survey conducted by~\cite{zhao2023survey}.
\begin{figure*}[t]
\centering
\setlength{\abovecaptionskip}{2pt}
\includegraphics[width=\linewidth]{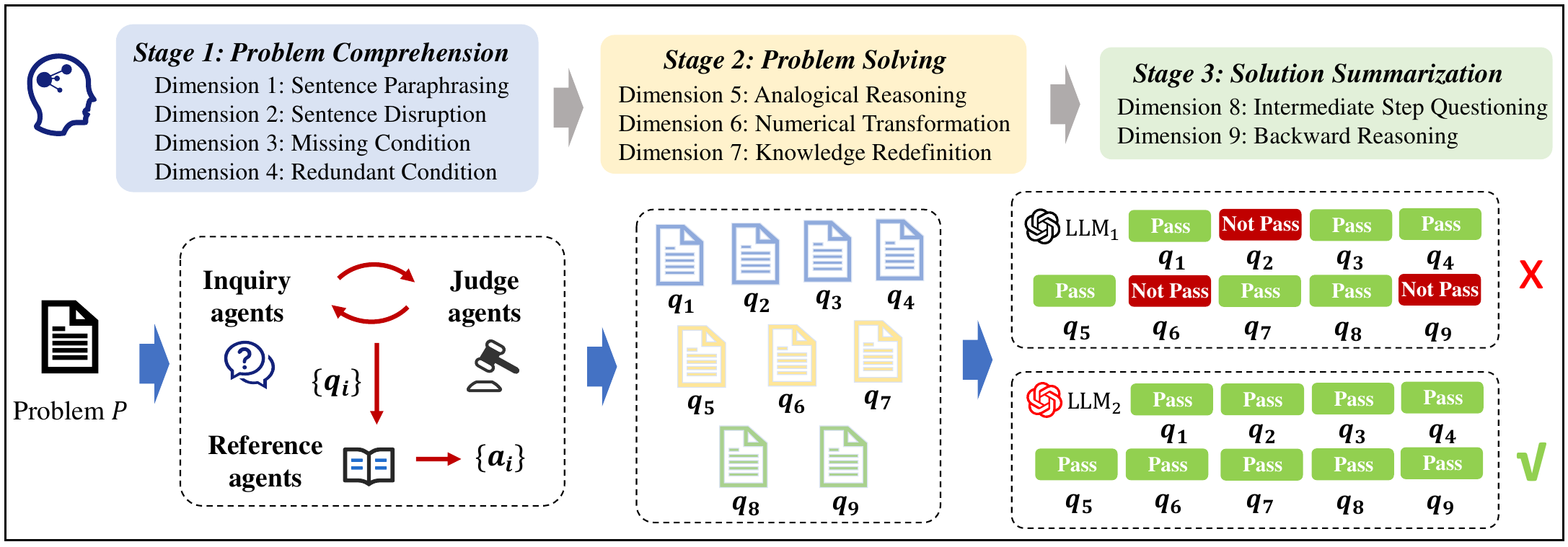}
\caption{Illustration of our CogMath framework.}
\label{figure_framework}
\end{figure*}
\subsection{Evaluation on LLM Mathematical Ability}
We categorize existing mathematical benchmarks from two perspectives: problem difficulty and problem types. In terms of difficulty, MATH~\citep{hendrycks2021measuring} and CHAMP~\citep{mao2024champ} are representative high school competition-level datasets, while GSM8K~\citep{cobbe2021training} and MAWPS~\citep{koncel2016mawps} are elementary-level math word problems. From the perspective of problem types, E-GSM~\citep{xu2024can} includes four categories of math problems of varying lengths to evaluate LLMs' generalization on longer contexts, TheoremQA~\citep{chen2023theoremqa} and MathBench~\citep{liu2024mathbench} test LLMs' ability to prove and apply theorems, while MathVista~\citep{lumathvista} and GeoEval~\citep{zhang2024geoeval} focus on visual reasoning and geometric reasoning. To mitigate the impact of data contamination, some studies introduce perturbations into existing benchmarks, such as GSM-1k~\cite{zhang2024careful}, GSM-Plus~\cite{li2024gsm}, and MPA~\citep{zhu2024dynamic}, which consist of manually-annotated/eight/five variations of GSM8K, respectively. 

However, these works lack in-depth exploration of models' reasoning processes, instead relying on a coarse overall accuracy metric. This makes it difficult to precisely identify at which cognitive stage the LLM encounters issues and to provide further guidance for improving LLMs.
\section{CogMath}
To achieve a comprehensive evaluation, we draw inspiration from how humans solve mathematical problems. Specifically, psychological theories indicate that human reasoning process consists of three stages: \emph{problem comprehension}, \emph{problem solving}, and \emph{solution summarization}~\citep{schoenfeld2014mathematical,lesh2003foundations,dehaene1999sources}. They build upon each other, with each stage taking the output of the previous one as input. \emph{Problem comprehension} involves analyzing problem $P$'s information, such as word semantics, text structure, and given conditions. \emph{Problem solving} stage combines the problem information with relevant knowledge to infer a solution. Finally, in \emph{solution summarization} stage, humans engage in self-summarization, reviewing their thought processes, organizing logical steps, and forming a structured methodology. 

Therefore, as illustrated in Figure~\ref{figure_framework}, in our \textbf{CogMath} framework, we evaluate LLMs' mathematical abilities from the above three stages. For each stage, we design multiple dimensions to assess LLMs from various perspectives. For instance, to assess a model's \emph{problem comprehension} stage, beyond investigating its performance after rephrasing the original problem, we can explore its sensitivity to changes in problem conditions, such as adding irrelevant information or disrupting problem sentences. Overall, for these three stages, we develop a total of nine dimensions that form a cohesive and comprehensive evaluation, with an example presented in Appendix~\ref{cogmath_details}. 

To quantify an LLM's performance in each dimension $i$, we design an ``\emph{Inquiry}-\emph{Judge}-\emph{Reference}'' multi-agent system. As shown in Figure~\ref{figure_framework}, the \emph{Inquiry} agent poses an inquiry $q_i$ based on $P$ that aligns with dimension $i$. The \emph{Judge} agent evaluates the quality of $q_i$ and repeatedly invokes the \emph{Inquiry} agent until a reasonable inquiry is obtained or the maximum number of iterations $\delta$ is reached. The \emph{Reference} agent generates an answer $a_i$ to $q_i$, which is used to assess whether a real LLM's response to $q_i$ is correct. Prompts for all agents are presented in Appendix~\ref{append_prompts}. For humans, truly mastering a mathematical problem requires a solid performance at each dimension. Hence, in CogMath, only when an LLM passes all dimensions can we conclude that it has genuinely mastered the problem $P$. Notably, these evaluation results also serve as a multifaceted analysis of the model, revealing gaps between its performance in each dimension and human cognition.
\subsection{Stage 1: \emph{Problem Comprehension}}
The \emph{problem comprehension} stage involves capturing the details of words, phrases, and sentences in the problem, as well as translating the mathematical conditions on a broader scale. To assess an LLM in this stage, we design four dimensions from fine-grained to coarse-grained granularities:

$\bullet$ \textbf{Dimension 1: Sentence Paraphrasing.} If a human truly understands a mathematical problem, she will demonstrate a robust understanding of the problem's meaning despite changes in wording or sentence structure. Inspired by this, this dimension evaluates the LLM's ability to understand a problem that has been rephrased using synonymous expressions. To achieve this, we ask the \emph{Inquiry} agent to pose a paraphrased version of the original problem $P$ as $q_1$, while preserving the mathematical essence (e.g., ``Jacob had \$21 ...'' in Table~\ref{cogmath_summarize}). Since the answer to the rephrased problem remains the same as the original, the \emph{Reference} agent can directly use the original answer as the reference $a_1$. 

$\bullet$ \textbf{Dimension 2: Sentence Disruption.} To prevent an LLM from simply memorizing a solution, we propose this dimension from a counterfactual perspective: the \emph{Inquiry} agent randomly disrupts the word order within each clause of the original problem, creating a ``pseudo problem'' $q_2$, where the words remain the same as in $P$, but from a human perspective, $q_2$ is unreadable and unsolvable. In this case, the \emph{Reference} agent does not need to generate an answer, as the expected response is simply ``unsolvable'', and the \emph{Judge} agent is also no longer required to make any judgments. If an LLM's response to $q_2$ is the same with the original answer, it indicates that this model is likely recalling an answer based on certain keywords or patterns rather than truly understanding the problem. Therefore, this dimension helps us assess whether the LLM is genuinely solving the problem or relying on superficial clues~\citep{sun2023text}. 

$\bullet$ \textbf{Dimension 3: Missing Condition.} For humans, understanding what the given conditions are in a math problem is a critical step in the comprehension process. If essential conditions are missing, we can recognize that the problem becomes unsolvable. Therefore, in this dimension, we still adopt a counterfactual approach: if, after the removal of a necessary condition, the LLM is still able to produce the original answer, it suggests that the model is relying on the semantic similarity to the memorized problem to map out the solution, rather than genuinely solving it. As illustrated in Appendix~\ref{append_prompts}.2.1, we ask the \emph{Inquiry} agent to omit one key condition from the original problem, presenting an underspecified version of $P$ as inquiry $q_3$. The \emph{Judge} agent needs to carefully assess whether only one condition has been removed and whether the inquiry $q_3$ does not alter any other parts of the original problem's formulation. 

$\bullet$ \textbf{Dimension 4: Redundant Condition.} In contrast to Missing Condition, this dimension introduces irrelevant conditions into the problem. For example, an extra condition such as ``Before meeting ... make sure it was correct'' might be added to the problem shown in Table~\ref{cogmath_summarize}. An LLM that truly masters problem $P$ should distinguish between essential and non-essential information, ensuring that unnecessary data does not interfere with the reasoning process. Therefore, the \emph{Inquiry} agent presents a problem $q_4$ with one redundant condition. The \emph{Judge} agent evaluates whether the extraneous detail does not affect the solution to the original problem, and the \emph{Reference} agent provides the original answer, as the added information should not affect the solution. 
\subsection{Stage 2: \emph{Problem Solving}}
Problem solving requires three orthogonal components: solving strategy, numerical calculation, and mathematical knowledge~\citep{sweller1988cognitive,jonassen2000toward}. The solving strategy is a logical organization specific to the problem, numerical calculation refers to arithmetic operations, and mathematical knowledge reflects common principles that apply across problems. To evaluate whether an LLM genuinely grasps them, we design the following three dimensions:

$\bullet$ \textbf{Dimension 5: Analogical Reasoning.} The solving strategy serves as a commonality across different problems, allowing a human to solve multiple similar problems using the same underlying logic. In this dimension, the \emph{Inquiry} agent presents a new problem that is conceptually consistent to, but not identical to, the original problem $P$, as $q_5$ (e.g., ``Tom had 21 comic ...'' in Table~\ref{cogmath_summarize}). This tests the LLM's ability to generalize the solving strategy, reflecting its grasp of the underlying reasoning thought. To be notice, $q_5$ preserves the original problem-solving process, maintaining the same approach, difficulty, and required knowledge, with no change to the core principles or complexity. 

$\bullet$ \textbf{Dimension 6: Numerical Transformation.} Generally, the solving strategy represents the essential structure of solution, and the final solving step can be seen as plugging the numerical values from the original problem into the strategy. Therefore, if a human has mastered the problem, changing the numerical values will not affect the ability to solve it~\cite{polya2014solve,leighton2007cognitive}. Based on this idea, in this dimension, the inquiry $q_6$ is a variant of the original problem $P$ that modifies its numerical values (e.g., replace the numbers ``21'' and ``100'' with ``30'' and ``120'' in Table~\ref{cogmath_summarize}). Since $q_6$ is a new problem, we instruct the \emph{Reference} agent to refer to the original answer and provide a corresponding new answer.

$\bullet$ \textbf{Dimension 7: Knowledge Redefinition.} Knowledge forms the foundation of human cognition, guiding how abstract principles are applied during the solution process~\citep{goldman1986epistemology,habermas2015knowledge,liu2023guiding,liu2025knowledge}. For example, solving a geometric problem may require the formula knowledge for triangle area. This understanding is flexible—if the problem redefines the formula of ``triangle area'', a human who truly grasps the concept will adapt her reasoning to fit the new definition. 

To assess if an LLM can do this, the \emph{Inquiry} agent modifies a key mathematical definition within problem $P$ by introducing a statement like ``Assume the area formula of a triangle is defined as four times the sum of the lengths of its sides'' in inquiry $q_7$. This redefinition forces the LLM to adapt its solution based on the modified concept. The \emph{Reference} agent then generates a new solution based on the redefined knowledge, and the \emph{Judge} agent assesses whether $q_7$ with the new definition is solvable.
\begin{table*}[t]
\small
\centering
\setlength{\abovecaptionskip}{5pt}
\setlength{\tabcolsep}{5pt} 
\begin{tabular}{@{\extracolsep{0pt}}c|c|cccccccc|c|c}
\toprule[1.5pt]
\multicolumn{2}{l|}{} & \multicolumn{8}{c|}{\footnotesize \textbf{MATH}} & \multirow{2}{*}{\centering \footnotesize \textbf{GSM8K}} & \multirow{2}{*}{\centering \footnotesize \textbf{MExam}} \\
\cline{3-10} \multicolumn{2}{l|}{} & \footnotesize \textbf{Avg} & \footnotesize \textbf{Alg} & \footnotesize \textbf{Count} & \footnotesize \textbf{Geo} & \footnotesize \textbf{Itmd} & \footnotesize \textbf{Num} & \footnotesize \textbf{Pre-Alg} & \footnotesize \textbf{Pre-Cal} & & \\ 
\hline
\multirow{3}{*}{\scriptsize GPT-4} & \scriptsize Vanilla & 0.758 & 0.908 & 0.783 & 0.660 & 0.580 & 0.792 & 0.879 & 0.574 & 0.954 & 0.807 \\ 
& \scriptsize CogMath                          & 0.393 & 0.532 & 0.395 & 0.276 & 0.197 & 0.337 & 0.587 & 0.266 & 0.671 & 0.364   \\ 
& $\Delta$                                  &  \cellcolor[gray]{0.9}-0.365 & \cellcolor[gray]{0.9}-0.376 & \cellcolor[gray]{0.9}-0.388 & \cellcolor[gray]{0.9}-0.384 & \cellcolor[gray]{0.9}-0.383 & \cellcolor[gray]{0.9}-0.455 & \cellcolor[gray]{0.9}-0.292 & \cellcolor[gray]{0.9}-0.308 & \cellcolor[gray]{0.9}-0.283 & \cellcolor[gray]{0.9}-0.440 \\ 
\hline
\multirow{3}{*}{\scriptsize GPT-3.5} & \scriptsize Vanilla & 0.482 & 0.672 & 0.426 & 0.390 & 0.276 & 0.415 & 0.693 & 0.273 & 0.838 &  0.531 \\
& \scriptsize CogMath                             & 0.176 & 0.280 & 0.108 & 0.121 & 0.062 & 0.109 & 0.315 & 0.088 & 0.424 &  0.192  \\ 
& $\Delta$                                     & \cellcolor[gray]{0.9}-0.306 & \cellcolor[gray]{0.9}-0.392 & \cellcolor[gray]{0.9}-0.318 & \cellcolor[gray]{0.9}-0.269 & \cellcolor[gray]{0.9}-0.214 & \cellcolor[gray]{0.9}-0.306 & \cellcolor[gray]{0.9} -0.378 & \cellcolor[gray]{0.9} -0.185 & \cellcolor[gray]{0.9} -0.414 & \cellcolor[gray]{0.9} -0.339 \\ 
\hline
\multirow{3}{*}{\scriptsize Gemini-1.5} & \scriptsize Vanilla & 0.615 & 0.812 & 0.535 & 0.489 & 0.423 & 0.555 & 0.781 & 0.479 & 0.922 & 0.739 \\ 
&\scriptsize CogMath                              & 0.291 & 0.428 & 0.247 & 0.173 & 0.142 & 0.206 & 0.455 & 0.205 & 0.500 & 0.338 \\ 
& $\Delta$                                        & \cellcolor[gray]{0.9}-0.325 &\cellcolor[gray]{0.9}-0.385 & \cellcolor[gray]{0.9}-0.288 & \cellcolor[gray]{0.9}-0.316 & \cellcolor[gray]{0.9}-0.281 & \cellcolor[gray]{0.9}-0.349 & \cellcolor[gray]{0.9} -0.326 & \cellcolor[gray]{0.9} -0.274 & \cellcolor[gray]{0.9} -0.422 & \cellcolor[gray]{0.9} -0.401 \\ 
\hline
\multirow{3}{*}{\scriptsize Llama3-8B} & \scriptsize Vanilla & 0.336 & 0.458 & 0.258 & 0.217 & 0.194 & 0.267 & 0.540 & 0.222 & 0.826 & 0.455 \\ 
& \scriptsize CogMath                               & 0.056 & 0.081 & 0.044 & 0.025 & 0.016 & 0.024 & 0.123 & 0.020 & 0.342 & 0.096  \\ 
& $\Delta$                                       & \cellcolor[gray]{0.9}-0.280 & \cellcolor[gray]{0.9}-0.377 & \cellcolor[gray]{0.9}-0.214 & \cellcolor[gray]{0.9} -0.192 & \cellcolor[gray]{0.9} -0.178 & \cellcolor[gray]{0.9} -0.243 & \cellcolor[gray]{0.9} -0.417 & \cellcolor[gray]{0.9} -0.202 & \cellcolor[gray]{0.9} -0.484 & \cellcolor[gray]{0.9} -0.359 \\ 
\hline
\multirow{3}{*}{\scriptsize Llama2-13B} & \scriptsize Vanilla & 0.106 & 0.142 & 0.080 & 0.073 & 0.051 & 0.074 & 0.196 & 0.059 & 0.446 & 0.267 \\ 
& \scriptsize CogMath                                & 0.008 & 0.013 & 0.004 & 0.004 & 0.003 & 0.001 & 0.016 & 0.004 & 0.064 & 0.024 \\ 
& $\Delta$                                        & \cellcolor[gray]{0.9}-0.098 & \cellcolor[gray]{0.9}-0.129 & \cellcolor[gray]{0.9}-0.076 & \cellcolor[gray]{0.9} -0.069 & \cellcolor[gray]{0.9} -0.048 & \cellcolor[gray]{0.9} -0.073 & \cellcolor[gray]{0.9} -0.180 & \cellcolor[gray]{0.9} -0.055 & \cellcolor[gray]{0.9} -0.382 & \cellcolor[gray]{0.9} -0.243 \\ 
\hline
\multirow{3}{*}{\scriptsize Mixtral-8x7B} & \scriptsize Vanilla & 0.374 & 0.495 & 0.306 & 0.278 & 0.238 & 0.265 & 0.529 & 0.339 & 0.575 & 0.506 \\  
& \scriptsize CogMath                                  & 0.092 & 0.147 & 0.053 & 0.058 & 0.037 & 0.028 & 0.165 & 0.079 & 0.212 & 0.133  \\ 
& $\Delta$                                          & \cellcolor[gray]{0.9}-0.282 & \cellcolor[gray]{0.9}-0.348 & \cellcolor[gray]{0.9}-0.253 & \cellcolor[gray]{0.9} -0.220 & \cellcolor[gray]{0.9} -0.201 & \cellcolor[gray]{0.9} -0.237 & \cellcolor[gray]{0.9} -0.364 & \cellcolor[gray]{0.9} -0.260 & \cellcolor[gray]{0.9} -0.363& \cellcolor[gray]{0.9}-0.373 \\ 
\hline
\multirow{3}{*}{\scriptsize DeepSeek-V2.5} & \scriptsize Vanilla & 0.747 & 0.915 & 0.730 & 0.597 & 0.548 & 0.780 & 0.870 & 0.625 & 0.951 & 0.855 \\  
& \scriptsize CogMath                                   & 0.368 & 0.519 & 0.346 & 0.284 & 0.207 & 0.285 & 0.526 & 0.233 & 0.646 & 0.342  \\ 
& $\Delta$                                           & \cellcolor[gray]{0.9}-0.379 & \cellcolor[gray]{0.9}-0.396 & \cellcolor[gray]{0.9}-0.384 & \cellcolor[gray]{0.9} -0.313 & \cellcolor[gray]{0.9} -0.341 & \cellcolor[gray]{0.9} -0.495 & \cellcolor[gray]{0.9} -0.344 & \cellcolor[gray]{0.9} -0.392 & \cellcolor[gray]{0.9} -0.305 & \cellcolor[gray]{0.9} -0.513 \\ 
\bottomrule[1.5pt]
\end{tabular}
\caption{Performance of different LLMs on vanilla datasets and our CogMath framework.}\label{performance_all}
\end{table*}
\subsection{Stage 3: \emph{Solution Summarization}}
After completing a problem-solving stage, humans often reflect on their reasoning process, summarizing the steps and the methodology behind their approach~\citep{cottrell2023critical,dewey2022we}. This summarization helps consolidate the understanding of not just the solution, but also the overall thought process, which can then be applied to similar problems in the future. In this stage, a human that truly masters the problem can accurately recall intermediate reasoning steps and verify the solution by working backward. To mimic these processes, we examine two critical dimensions:

$\bullet$ \textbf{Dimension 8: Intermediate Step Questioning.} In human reasoning, breaking down the problem-solving process into smaller, manageable steps is essential for clarity and learning. Beyond evaluating the final answer, assessing whether an LLM has precisely understood the intermediate steps is an indispensable part of determining if it truly grasps the solution. Therefore, in this dimension, the \emph{Inquiry} agent presents an inquiry $q_8$ that asks an LLM to explain one of the key intermediate steps during the problem-solving process (e.g., step 2 in Appendix~\ref{append_prompts}.7.1). This ensures that the model is not just arriving at a correct final answer by coincidence or pattern recognition, but is following a clear, logical sequence throughout the entire solution. Then, the \emph{Judge} agent checks whether $q_8$ corresponds to a specific step, and the \emph{Reference} agent generates an explanation for this step based on the original solution.

$\bullet$ \textbf{Dimension 9: Backward Reasoning.} Inspired by~\cite{yumetamath,weng2023large}, backward reasoning is a crucial and challenging task. It refers to inferring missing information from the solution, mirroring how humans check their thought by retracing their reasoning to ensure there are no mistakes~\citep{rips1994psychology}. Therefore, it can be used to evaluate whether LLMs maintain consistency and logical coherence from both directions—forward and backward. If a model truly understands the problem-solving process, it should be able to  perform this reverse reasoning without contradictions. 

For this purpose, our \emph{Inquiry} agent formulates inquiry $q_9$ by masking a key numerical value from the original problem $P$ and requiring the model to infer the missing value based on the original solution. The \emph{Reference} agent directly takes the masked value as the answer $a_9$, and the \emph{Judge} agent evaluates whether the masked problem, when combined with the original answer, remains solvable.
\begin{table*}[t]
\small
\centering
\setlength{\abovecaptionskip}{2pt}
\setlength{\tabcolsep}{5.5pt} 
\begin{tabular}{c|c|cccccccc|c|c}
\toprule[1.5pt]
\multicolumn{2}{l|}{}    & \multicolumn{8}{c|}{\footnotesize \textbf{MATH}} & \multirow{2}{*}{\centering \footnotesize \textbf{GSM8K}} & \multirow{2}{*}{\footnotesize \textbf{MExam}} \\
\cline{3-10} \multicolumn{2}{l|}{} & \footnotesize \textbf{Avg} & \footnotesize \textbf{Alg} & \footnotesize \textbf{Count} & \footnotesize \textbf{Geo} & \footnotesize \textbf{Itmd} & \footnotesize \textbf{Num} & \footnotesize \textbf{Pre-Alg} & \footnotesize \textbf{Pre-Cal} & & \\ 
\hline
\multirow{3}{*}{\scriptsize GPT-4} & \scriptsize \emph{Stage} 1 & 0.630 & 0.813 & 0.671 & 0.459 & 0.401 & 0.635 & 0.798 & 0.452 & 0.851 &  0.690\\ 
&\scriptsize \emph{Stage} 2                        & \textbf{0.532} & \textbf{0.683} & \textbf{0.534} & \textbf{0.395} & \textbf{0.323} & \textbf{0.485} & \textbf{0.728} & \textbf{0.401} &0.870& 0.624 \\ 
&\scriptsize \emph{Stage} 3                        & 0.699 & 0.773 & 0.698 & 0.595 & 0.604 & 0.711 & 0.790 & 0.630 & \textbf{0.832} & \textbf{0.600} \\ 
\hline
\multirow{3}{*}{\scriptsize GPT-3.5} &\scriptsize \emph{Stage} 1 & 0.359 & 0.561  & 0.283 & 0.246 & \textbf{0.147} & 0.257 & 0.571 & \textbf{0.194} & \textbf{0.636} & 0.443 \\
&\scriptsize \emph{Stage} 2                          & \textbf{0.334} & \textbf{0.482}  & \textbf{0.262} & \textbf{0.228} & 0.161 & \textbf{0.250} & \textbf{0.543} & 0.209 & 0.707 & \textbf{0.407}     \\ 
&\scriptsize \emph{Stage} 3                          & 0.486 & 0.574  & 0.460 & 0.397 & 0.396 & 0.465 & 0.563 & 0.443 & 0.662 & 0.474      \\ 
\hline
\multirow{3}{*}{\scriptsize Gemini-1.5} &\scriptsize \emph{Stage} 1 & 0.509 & 0.715 & 0.428 & 0.388 & 0.307 & 0.415 & 0.692 & 0.372 & 0.829 & 0.618 \\ 
&\scriptsize \emph{Stage} 2                             & \textbf{0.421} & \textbf{0.586} & \textbf{0.380} & \textbf{0.284} & \textbf{0.240} & \textbf{0.300} & \textbf{0.629} & \textbf{0.302} & 0.806 & \textbf{0.579}  \\ 
&\scriptsize \emph{Stage} 3                             & 0.659 & 0.741 & 0.660 & 0.534 & 0.571 & 0.678 & 0.718 & 0.623 & \textbf{0.748} & 0.653   \\ 
\hline
\multirow{3}{*}{\scriptsize Llama3-8B} &\scriptsize \emph{Stage} 1 & 0.168 & 0.256 & 0.133 & 0.094 & \textbf{0.059} & \textbf{0.094} & 0.318 & \textbf{0.090} & 0.607 &  0.301 \\ 
&\scriptsize \emph{Stage} 2                            & \textbf{0.160} & \textbf{0.215} & \textbf{0.118} & \textbf{0.079} & 0.079 & 0.106 & \textbf{0.307} & 0.104 & 0.626 &  \textbf{0.294}        \\ 
&\scriptsize \emph{Stage} 3                            & 0.303 & 0.356 & 0.314 & 0.240 & 0.235 & 0.244 & 0.392 & 0.267 & \textbf{0.556} & 0.348          \\ 
\hline
\multirow{3}{*}{\scriptsize Llama2-13B} &\scriptsize \emph{Stage} 1 & \textbf{0.039} & 0.063 & 0.076 & \textbf{0.019} & \textbf{0.019} & \textbf{0.012} & 0.085 & \textbf{0.011} & 0.243 &  \textbf{0.118} \\ 
&\scriptsize \emph{Stage} 2                             & 0.047 & \textbf{0.062} & \textbf{0.027} & 0.029 & 0.037 & 0.024 & \textbf{0.080} & 0.037 & 0.253 &  0.133      \\ 
&\scriptsize \emph{Stage} 3                             & 0.117 & 0.132 & 0.122 & 0.081 & 0.113 & 0.094 & 0.140 & 0.103 & \textbf{0.232} &  0.289         \\ 
\hline
\multirow{3}{*}{\scriptsize Mixtral-8x7B} &\scriptsize \emph{Stage} 1 & \textbf{0.200} & \textbf{0.308} & \textbf{0.131} & \textbf{0.127} & \textbf{0.094} & \textbf{0.113} & \textbf{0.327} & \textbf{0.150} & \textbf{0.400} &  0.364 \\  
&\scriptsize \emph{Stage} 2                               & 0.224 & 0.328 & 0.139 & 0.146 & 0.136 & 0.133 & 0.344 & 0.185 & 0.430 & \textbf{0.332}  \\ 
&\scriptsize \emph{Stage} 3                               & 0.398 & 0.434 & 0.376 & 0.372 & 0.341 & 0.337 & 0.490 & 0.374 & 0.569 & 0.432  \\ 
\hline
\multirow{3}{*}{\scriptsize DeepSeek-V2.5} &\scriptsize \emph{Stage} 1 & 0.649 & 0.844 & 0.578 & 0.507 & 0.455 & 0.683 & 0.780 & 0.491 & 0.832 & 0.717 \\  
&\scriptsize \emph{Stage} 2                                & \textbf{0.526} & \textbf{0.695} & \textbf{0.496} & \textbf{0.411} & \textbf{0.328} & \textbf{0.463} & \textbf{0.723} & \textbf{0.357} & 0.850 &  0.672         \\ 
&\scriptsize \emph{Stage} 3                                & 0.681 & 0.762 & 0.692 & 0.610 & 0.607 & 0.644 & 0.741 & 0.623 & \textbf{0.817} &  \textbf{0.541}         \\ 
\bottomrule[1.5pt]
\end{tabular}
\caption{Performance of different LLMs at each cognitive stage.}\label{ablation}
\end{table*}
\section{Evaluation}
\subsection{Experimental Setup}
We evaluated seven mainstream LLMs, including four closed-source models: GPT-4~\citep{achiam2023gpt}, GPT-3.5-Turbo~\citep{chatgpt2023}, Gemini-1.5-Flash~\citep{team2023gemini}, and DeepSeek-V2.5~\cite{liu2024deepseek}, as well as three open-source models: Llama3-8B~\citep{meta2024introducing}, Llama2-13B~\citep{touvron2023llama}, and Mixtral-8x7B-Instruct~\citep{mistral2023mixtral}. The implementation details and problem sets of CogMath are described in Appendix~\ref{implementation_details}. We adopt \emph{Pass Rate} (PR) as our metric. This is because, in CogMath, dimensions 2 and 3 are based on counterfactual settings. Therefore, for inquiries $q_2$ and $q_3$, the expected response is ``unsolvable'' (as shown in Table~\ref{cogmath_summarize}), and when the LLM's response differs from the original answer, we consider it to have passed the corresponding inquiry. For the remaining seven dimensions and the original dataset, \emph{Pass} refers to correctly answer. 
\subsection{Main Results}\label{exp_main}
Table~\ref{performance_all} presents the original results (``Vanilla'') of all LLMs as well as their performance under our CogMath framework. First, there is a significant decrease of 30\%-40\% in pass rates for all models, indicating that the mathematical abilities they display on public benchmarks may not be as genuine and reliable as they appear. Even GPT-4 successfully passes only 39.3\% and 67.1\% of problems in MATH and GSM8K datasets, respectively. Second, on the more challenging MATH dataset, the most powerful models (i.e., perform best in ``Vanilla''), GPT-4 and DeepSeek-V2.5, exhibit the largest drops, with $\Delta=$36.5\% and 37.9\%, respectively. However, on the simpler GSM8K dataset, their declines are the smallest, with $\Delta=$28.3\% and 30.5\%, respectively. This suggests that the extent to which the capabilities of LLMs are overestimated does not diminish as the models become stronger, but rather remains a widespread phenomenon unrelated to model size or dataset difficulty. Third, the issue of overestimated model capability persists on our newly constructed MExam dataset, which has not been used for training these LLMs. On one hand, this suggests that the overestimation is not solely due to data contamination. On the other hand, this phenomenon demonstrates that simply introducing more test problems may be insufficient to assess the true mathematical abilities of LLMs.
\begin{figure*}[t]
  \centering
  \setlength{\abovecaptionskip}{2pt}
  \begin{subfigure}[b]{0.25\textwidth}
    \includegraphics[width=\textwidth]{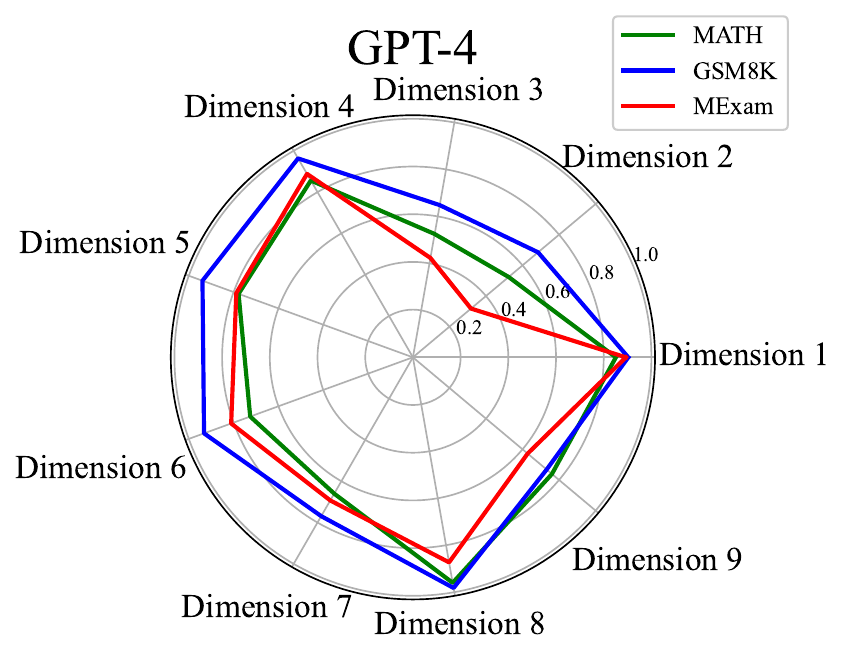}
  \end{subfigure}
  \hspace{-0.1in}
  \begin{subfigure}[b]{0.25\textwidth}
    \includegraphics[width=\textwidth]{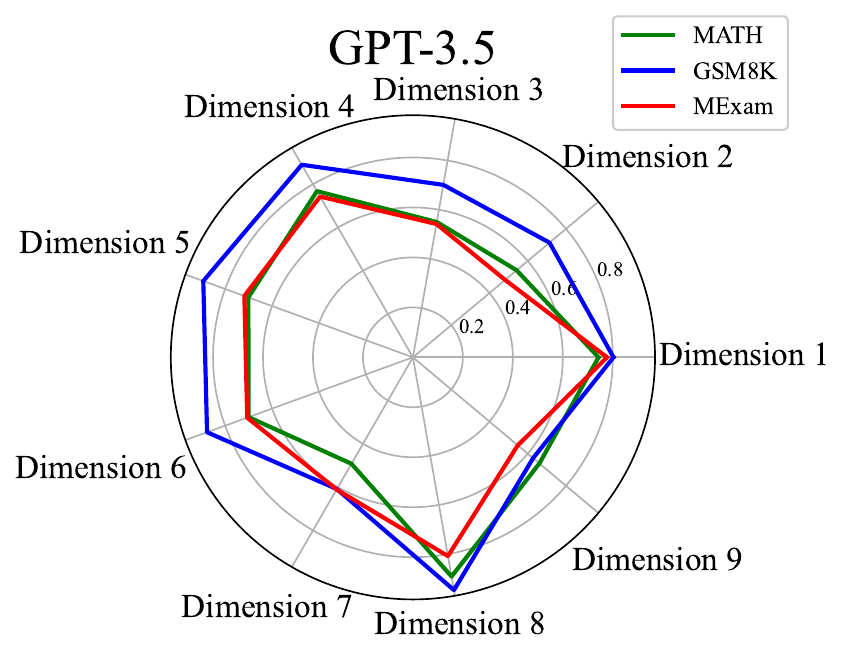}
  \end{subfigure}
  \hspace{-0.1in}
  \begin{subfigure}[b]{0.25\textwidth}
    \includegraphics[width=\textwidth]{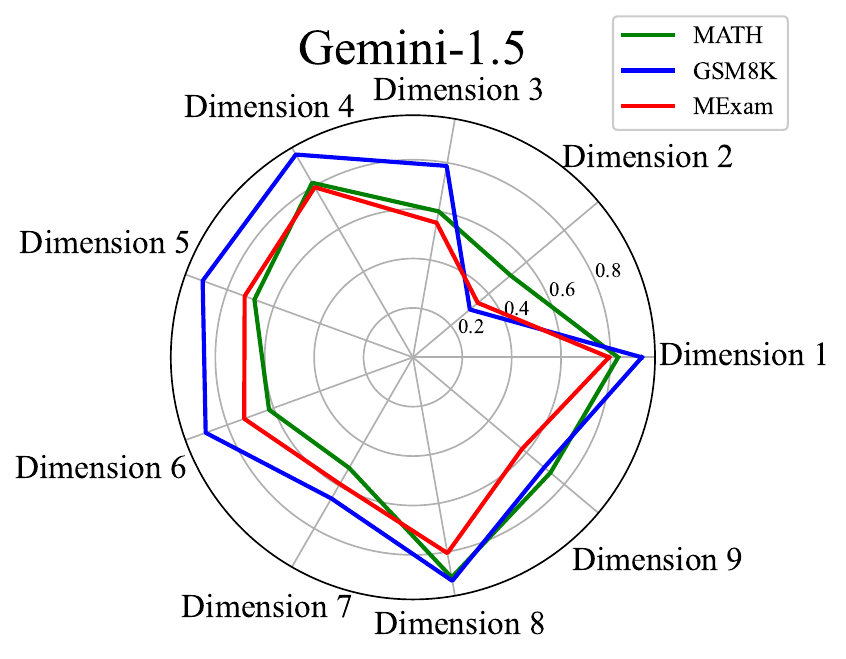}
  \end{subfigure}
  \hspace{-0.1in}
  \begin{subfigure}[b]{0.25\textwidth}
    \includegraphics[width=\textwidth]{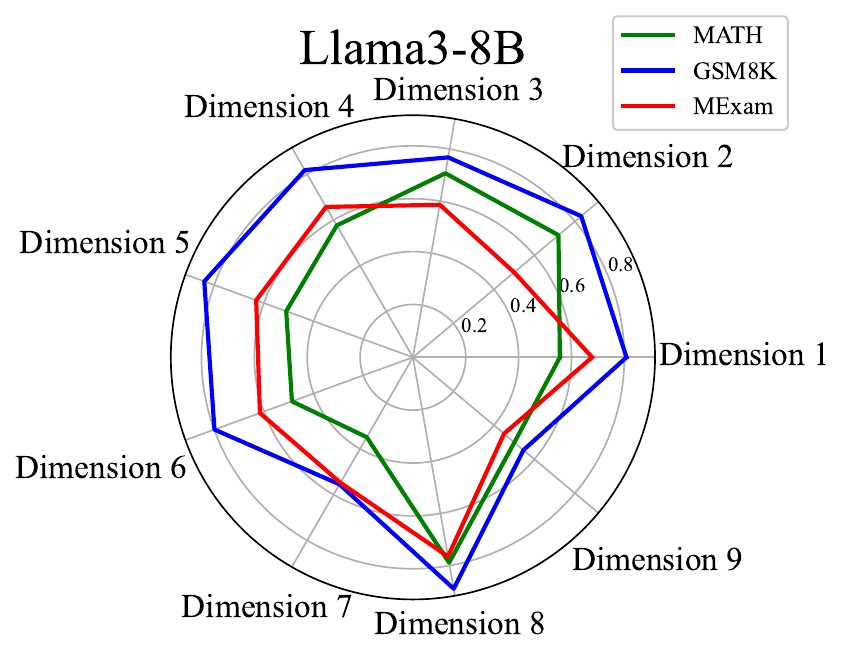}
  \end{subfigure}
  \hspace{-0.1in}
  \begin{subfigure}[b]{0.25\textwidth}
    \includegraphics[width=\textwidth]{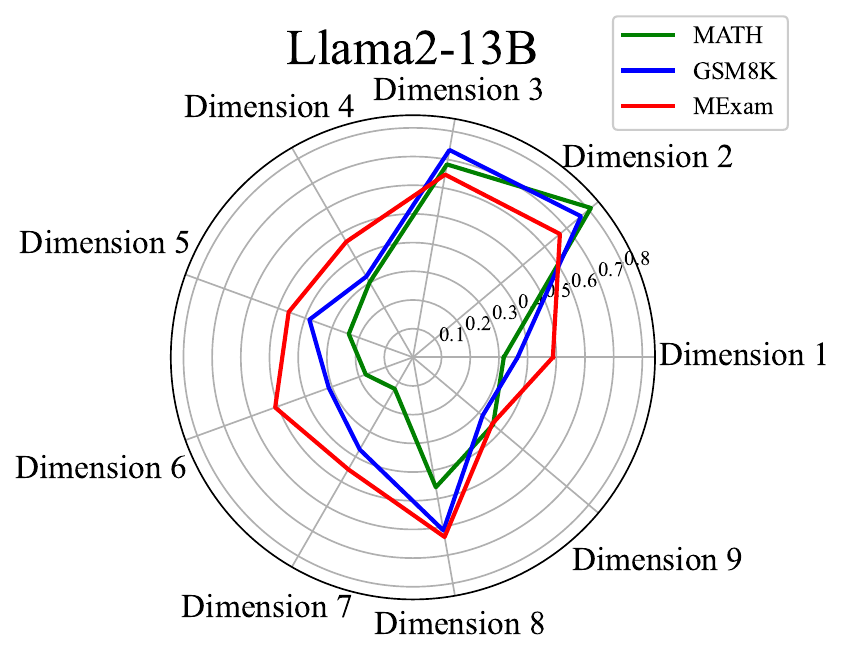}
  \end{subfigure}
  \hspace{-0.1in}
  \begin{subfigure}[b]{0.25\textwidth}
    \includegraphics[width=\textwidth]{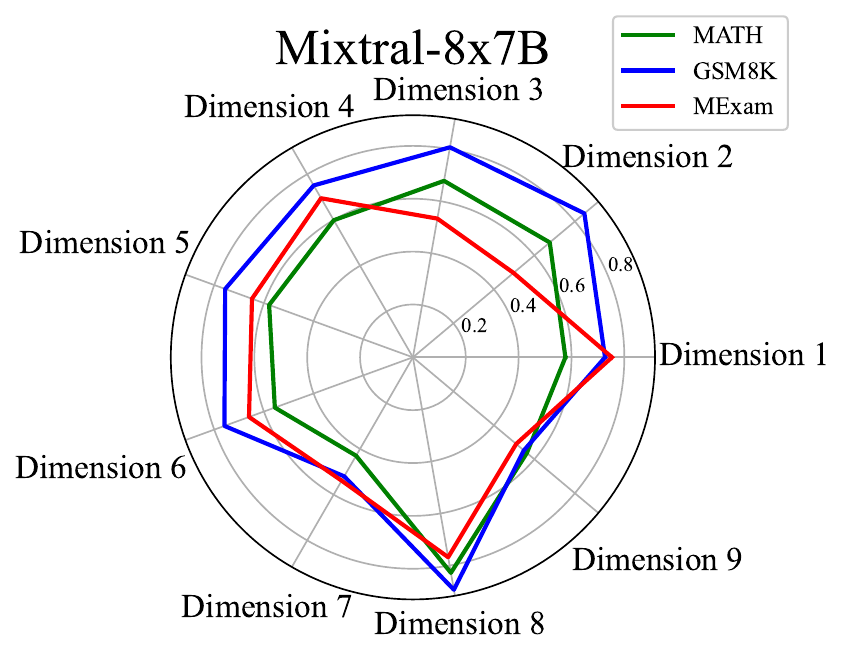}
  \end{subfigure}
  \hspace{-0.1in}
  \begin{subfigure}[b]{0.25\textwidth}
    \includegraphics[width=\textwidth]{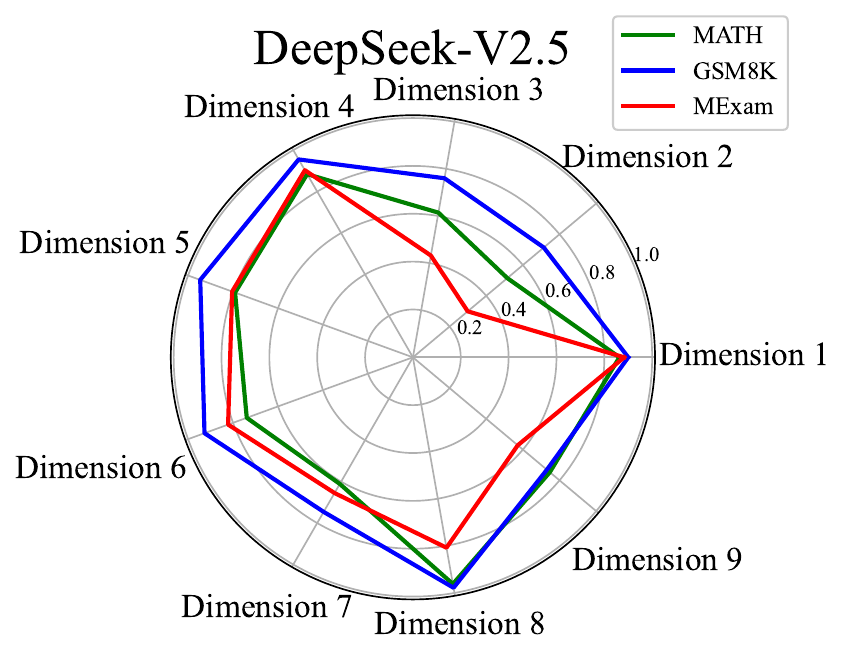}
  \end{subfigure}
  \caption{\emph{Relative Pass Rate} (RPR) of different LLMs in each dimension.}
  \label{fig:dimension}
\end{figure*}
\subsection{Analysis on Three Cognitive Stages}\label{section_cog_stages}
To further analyze the extent to which LLMs grasp different cognitive stages, we present the \emph{Pass Rate} at different stages in Table~\ref{ablation}, with the stage having the lowest pass rate highlighted in bold. Specifically, we first observe that for weaker LLMs (e.g., Llama2-13B), their pass rates in Stage 1 (i.e., \emph{problem comprehension}) are the lowest, indicating that these models already exhibit deficiencies in fundamental understanding. For more advanced models (e.g., GPT-4, DeepSeek-V2.5), their comprehension abilities appear more stable. However, they struggle significantly with mastering Stage 2 (e.g., GPT-4 and Deepseek exhibit pass rates of only 53.2\% and 52.6\% on MATH, respectively). In Section~\ref{section_cog_dimension}, we reveal that the main reason is that their grasp of knowledge is still unstable. Finally, the pass rate in Stage 3 remains below $0.85$. This suggests that current LLMs are more suited for forward reasoning, i.e., generating answers based on the problems, but struggle to assess whether the solution aligns with the original problem from a backward perspective. This finding is consistent with existing research that shows LLMs may find it challenging to verify the correctness of their own answers~\citep{huanglarge}.
\subsection{Analysis on Nine Cognitive Dimensions}\label{section_cog_dimension}
Furthermore, we analyze the performance of LLMs in each dimension. Specifically, for dimension $i$, we calculate a \emph{Relative Pass Rate} (RPR) defined as: RPR = $\frac{|Pass_i\cap Pass|}{|Pass|}$. Here, $Pass_i$ denotes the problems where the LLM successfully passes their corresponding inquiry $q_i$, and $Pass$ refers to the problems correctly answered by this model. It is important to note that a higher RPR indicates better robustness and stability of the LLM's capabilities in that dimension. This is because the model's performance on corresponding inquiries is highly consistent with its performance on the original problems, making it less likely to exhibit defects when it answers the original problem correctly. Conversely, a lower RPR signifies a more detrimental impact on LLM performance, suggesting that the model exhibits lower adaptability to that type of inquiry. 

Overall, from Figure~\ref{fig:dimension}, DeepSeek-V2.5 and GPT-4 exhibit the most balanced performance across multiple dimensions, followed by GPT-3.5, Mixtral-8x7B, Gemini-1.5, and Llama3-8B, with Llama2-13B performing the worst. Secondly, regarding the four dimensions in \emph{problem comprehension} stage, an important observation is that GPT-4, GPT-3.5, Gemini-1.5, and DeepSeek-V2.5 underperform in Dimensions 2 and 3, even lagging behind Llama2-13B and Llama3-8B. We speculate that this is because most training data for current LLMs is composed of solvable math problems. After being trained on such data, when facing an unsolvable problem, current LLMs may inherently ``over-correct'' the problem into a solvable one, aligning it more closely with their training data. This insight suggests that in order to equip LLMs with more human-like cognitive capabilities, it is necessary to cultivate critical thinking skills rather than mere imitation of training data. Thirdly, for the three dimensions associated with \emph{problem solving} stage, Dimension 7 accounts for the low pass rate discussed in Section~\ref{section_cog_stages}. This indicates that current LLMs treat knowledge more as rigid memorization and application, rather than integrating it organically and flexibly into the reasoning process. Lastly, in \emph{solution summarization} stage, nearly all LLMs demonstrate higher RPR values in Dimension 8, suggesting that they are quite adept at explaining reasoning steps. However, the performance in Dimension 9 indicates that these models struggle to use conclusions to reversely derive conditions, which explain why they are difficult to self-verify the correctness of their own answers.
\begin{table}[t]
    \small
    \setlength{\abovecaptionskip}{2pt}
    \renewcommand{\arraystretch}{1.05}
    \setlength{\tabcolsep}{1pt} 
    \centering
    \begin{tabular}{c|c|c|c|c}
    \toprule[1.5pt]
    \multicolumn{2}{l|}{} &  CogMath & CogMath(CoT) & CogMath(ICL) \\
    \hline
    \multirow{3}{*}{MATH} & GPT-4 & 0.393 & 0.380 & 0.368 \\
                         & GPT-3.5 & 0.176 & 0.169 & 0.167 \\
                         & Gemini-1.5 & 0.291 & 0.242 & 0.250 \\
    \hline
   \multirow{3}{*}{GSM8K} & GPT-4 & 0.671 & 0.680 & 0.676 \\
                         & GPT-3.5 & 0.424 & 0.442 & 0.466 \\
                         & Gemini-1.5 & 0.500 & 0.585 & 0.518 \\
   \bottomrule[1.5pt]
    \end{tabular}
    \caption{Performances of LLM enhancement methods.}
    \label{perform_prompt}
    \vspace{-15pt}
\end{table}
\subsection{Effect of LLM Enhancement Methods}\label{exp_prompt}
We explore the impact of two commonly used reasoning enhancement methods on LLMs' mathematical abilities: Chain-of-Thought (CoT)~\citep{wei2022chain} and In-Context Learning (ICL)~\citep{dong2022survey}. For CoT, we prompt the LLM to answer each inquiry in CogMath ``step by step''. For ICL, we adopt a one-shot setting where, for each dimension $i$, we randomly sample a problem $P_i$ from the training set and use CogMath to construct an (inquiry $q^i_P$, answer $a^i_P$) pair as the demonstration. 

From Table~\ref{perform_prompt}, these techniques led to a performance decrease of 0.7\% ($0.176\rightarrow 0.169$) to 4.9\% ($0.291\rightarrow 0.242$) on MATH but an increase of 0.5\% ($0.671\rightarrow 0.676$) to 8.5\% ($0.500\rightarrow 0.585$) on GSM8K. These results suggest that prompting techniques may not fundamentally enhance the mathematical abilities of LLMs. Instead, they serve more as an auxiliary tool, which can bring more positive effects on simpler datasets. In some cases, ICL might even limit the model's problem-solving flexibility. For more analyses of the effects at the dimension level, please refer to Appendix~\ref{dimension_llm}.
\subsection{Error Analysis}\label{exp_error}
From Sections~\ref{exp_main} to~\ref{section_cog_dimension}, we verify that the primary reason for LLMs making errors in our CogMath is due to their deficiencies in abilities corresponding to Dimensions 2, 3, 7, and 9. In this section, we further investigate how the characteristics of the problems influence LLMs' errors. Specifically, we take the MATH dataset as an example and explore the influence of problem difficulty and problem length. Problem difficulty is measured by the dataset's inherent ``level'' labels, which include five tiers. For problem length, we divide all problems into five levels using an equal-frequency binning approach.

\begin{figure}[t]
  \centering
  \setlength{\abovecaptionskip}{2pt}
  \begin{subfigure}[b]{0.49\textwidth}
    \includegraphics[width=\textwidth]{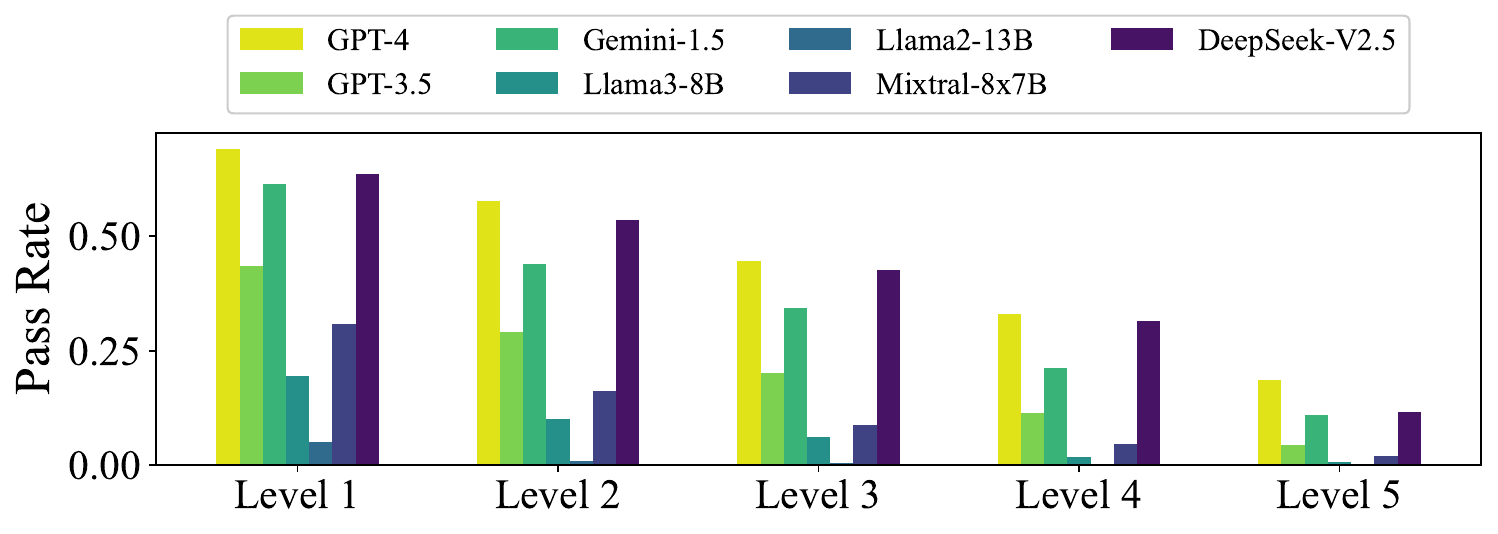}
  \end{subfigure}
  \hspace{-0.1in}
  \begin{subfigure}[b]{0.49\textwidth}
    \includegraphics[width=\textwidth]{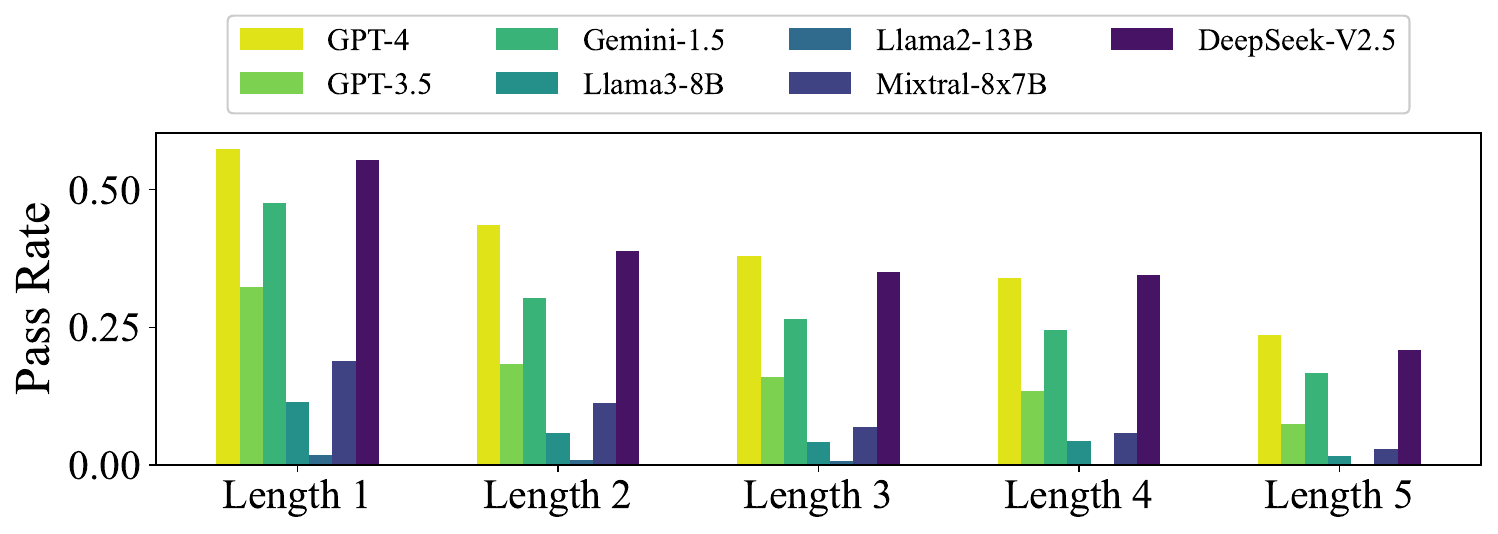}
  \end{subfigure}
  \caption{Relationship between LLM performance with problem characteristics (i.e., difficulty and length).}
  \label{fig:distribution}
\vspace{-12pt}
\end{figure}
From Figure~\ref{fig:distribution}, we can first observe that as problem difficulty increases, the performance of all LLMs declines significantly. More specifically, most models only perform well on level 1 problems, while only GPT-4 and DeepSeek-V2.5 demonstrate proficiency on more than half of the problems at both levels 1 and 2. Secondly, as problem length increases, the LLM performance also shows some decline, though it is less significant compared to the impact of problem difficulty. This suggests that problem length has a relatively lower correlation with model performance. Based on these observations, we think future improvements in LLMs' mathematical abilities could focus on enhancing their capacity to handle more complex problems, particularly those in higher difficulty levels.

\subsection{Human Verification of Agents}\label{human_evaluation}
In this section, we provide evidence to make sure all our agents faithfully finish their jobs. Specifically, for each \emph{Inquiry} agent, we have designed a \emph{Judge} agent to evaluate the quality of its output. If the \emph{Judge} agent determines that the output does not meet the required quality standards, we ask the \emph{Inquiry} agent to regenerate the response. 

To assess the effectiveness of the \emph{Judge} agent, we invite 5 well-trained annotators with undergraduate degrees to verify the final inquires approved by the \emph{Judge} agent, evaluating the rate at which they align with the intended dimensions. The evaluation template is same with the prompts for the \emph{Judge} agent as presented in Appendix~\ref{append_prompts}. The evaluation protocol was approved by the Ethics Review Board and all annotators have been informed about the intended use of the data. The results for each dimension on 500 randomly selected problems are shown in Table~\ref{eval_judge} (Since Dimension 2 relies on rule-based sentence disruption, there is no need for a \emph{Judge} agent).

Regarding the \emph{Reference} agents, Dimensions 1–4 simply use the original problem's answer as the correct response for $q_i$, while Dimension 9 automatically extracts the masked value from the \emph{Inquiry} agent's output as $a_9$. Here, we focus on evaluating Dimensions 5–8. We invite the same annotators to evaluate the results of another 500 randomly selected problems. As shown in Table~\ref{eval_reference}, the answers generated by the \emph{Reference} agent achieve a pass rate of 95\% across all dimensions, which ensures the quality of its outputs.
\begin{table}[t]
    \small
    \setlength{\abovecaptionskip}{4pt}
    \setlength{\tabcolsep}{2pt} 
    \centering
    \begin{tabular}{c|c|c|c|c|c|c|c|c}
    \toprule[1.5pt]
    Dimension &  D.1 & D.3 & D.4 & D.5 & D.6 & D.7 &  D.8 & D.9 \\
    \hline
    Rate (\emph{Judge}) & 0.984& 0.992& 0.964& 0.986& 0.986& 0.952& 0.990& 0.950\\
   \bottomrule[1.5pt]
    \end{tabular}
    \caption{Human Verification of \emph{Judge} agents.}
    \label{eval_judge}
    \vspace{-5pt}
\end{table}

\begin{table}[t]
    \small
    \setlength{\abovecaptionskip}{4pt}
    \setlength{\tabcolsep}{2pt} 
    \centering
    \begin{tabular}{c|c|c|c|c}
    \toprule[1.5pt]
    Dimension &  D.5 & D.6 & D.7 &  D.8 \\
    \hline
    Rate (\emph{Reference}) & 0.954 & 0.968 & 0.952 & 0.986 \\
   \bottomrule[1.5pt]
    \end{tabular}
    \caption{Human Verification of \emph{Reference} agents.}
    \label{eval_reference}
    \vspace{-5pt}
\end{table}
\section{Conclusion and Discussion}
In this paper, we introduced CogMath, a comprehensive and scientific evaluation framework that assesses the mathematical abilities of LLMs across three cognitive stages and nine dimensions of humans. The findings indicated that the abilities of current mainstream LLMs are overestimated by approximately 30\%-40\%. Moreover, we located the strength and weakness of different LLMs and verified that prompting techniques such as CoT and ICL do not genuinely enhance their mathematical proficiency. For future work, we discuss some valuable directions in Appendix~\ref{append_discuss}.

\section*{Impact Statement}
CogMath presents a valuable evaluation framework for assessing the reasoning capabilities of LLMs, but certain aspects warrant further consideration. Firstly, CogMath relies heavily on interactions among multiple LLM agents, which may limit the scalability due to the computational costs associated with generating inquiries for large-scale benchmarks. Secondly, while our study contributes to the understanding of LLM enhancement methods, it does not encompass all existing techniques, notably excluding widely adopted approaches such as Program of Thoughts (PoT) and Retrieval-Augmented Generation (RAG). These methods involve additional processes, such as code generation and information retrieval, which were beyond the scope of this work. Investigating whether these techniques can fundamentally enhance the reasoning abilities of LLMs remains an important direction for future research. 

\section*{Acknowledgement}
This research was partially supported by the National Natural Science Foundation of China (Grants No.62477044, 62406303, U23A20319), Anhui Provincial Natural Science Foundation (No. 2308085QF229), the Fundamental Research Funds for the Central Universities (No.WK2150110038). Zhenya Huang gratefully acknowledges the support of the Young Elite Scientists Sponsorship Program by CAST (No. 2024QNRC001).

\bibliography{reference}

\begin{thebibliography}{56}
\providecommand{\natexlab}[1]{#1}
\providecommand{\url}[1]{\texttt{#1}}
\expandafter\ifx\csname urlstyle\endcsname\relax
  \providecommand{\doi}[1]{doi: #1}\else
  \providecommand{\doi}{doi: \begingroup \urlstyle{rm}\Url}\fi

\bibitem[Achiam et~al.(2023)Achiam, Adler, Agarwal, Ahmad, Akkaya, et~al.]{achiam2023gpt}
Achiam, J., Adler, S., Agarwal, S., Ahmad, L., Akkaya, I., et~al.
\newblock Gpt-4 technical report.
\newblock \emph{arXiv preprint arXiv:2303.08774}, 2023.

\bibitem[Besta et~al.(2024)Besta, Blach, Kubicek, Gerstenberger, et~al.]{besta2024graph}
Besta, M., Blach, N., Kubicek, A., Gerstenberger, R., et~al.
\newblock Graph of thoughts: Solving elaborate problems with large language models.
\newblock In \emph{Proceedings of the AAAI Conference on Artificial Intelligence}, volume~38, pp.\  17682--17690, 2024.

\bibitem[Chen et~al.(2024)Chen, Lin, Han, and Sun]{chen2024benchmarking}
Chen, J., Lin, H., Han, X., and Sun, L.
\newblock Benchmarking large language models in retrieval-augmented generation.
\newblock In \emph{Proceedings of the AAAI Conference on Artificial Intelligence}, volume~38, pp.\  17754--17762, 2024.

\bibitem[Chen et~al.(2023)Chen, Yin, Ku, Lu, et~al.]{chen2023theoremqa}
Chen, W., Yin, M., Ku, M., Lu, P., et~al.
\newblock Theoremqa: A theorem-driven question answering dataset.
\newblock In \emph{Proceedings of the 2023 Conference on Empirical Methods in Natural Language Processing}, pp.\  7889--7901, 2023.

\bibitem[Cobbe et~al.(2021)Cobbe, Kosaraju, Bavarian, et~al.]{cobbe2021training}
Cobbe, K., Kosaraju, V., Bavarian, M., et~al.
\newblock Training verifiers to solve math word problems.
\newblock \emph{arXiv preprint arXiv:2110.14168}, 2021.

\bibitem[Cottrell(2023)]{cottrell2023critical}
Cottrell, S.
\newblock \emph{Critical thinking skills: Effective analysis, argument and reflection}.
\newblock Bloomsbury Publishing, 2023.

\bibitem[Dehaene et~al.(1999)Dehaene, Spelke, Pinel, Stanescu, and Tsivkin]{dehaene1999sources}
Dehaene, S., Spelke, E., Pinel, P., Stanescu, R., and Tsivkin, S.
\newblock Sources of mathematical thinking: Behavioral and brain-imaging evidence.
\newblock \emph{Science}, 284\penalty0 (5416):\penalty0 970--974, 1999.

\bibitem[Dewey(2022)]{dewey2022we}
Dewey, J.
\newblock \emph{How we think}.
\newblock DigiCat, 2022.

\bibitem[Dong et~al.(2022)Dong, Li, Dai, Zheng, Wu, Chang, Sun, Xu, and Sui]{dong2022survey}
Dong, Q., Li, L., Dai, D., Zheng, C., Wu, Z., Chang, B., Sun, X., Xu, J., and Sui, Z.
\newblock A survey on in-context learning.
\newblock \emph{arXiv preprint arXiv:2301.00234}, 2022.

\bibitem[Goldman(1986)]{goldman1986epistemology}
Goldman, A.~I.
\newblock \emph{Epistemology and cognition}.
\newblock harvard university Press, 1986.

\bibitem[Guo et~al.(2025)Guo, Yang, Zhang, Song, Zhang, Xu, Zhu, Ma, Wang, Bi, et~al.]{guo2025deepseek}
Guo, D., Yang, D., Zhang, H., Song, J., Zhang, R., Xu, R., Zhu, Q., Ma, S., Wang, P., Bi, X., et~al.
\newblock Deepseek-r1: Incentivizing reasoning capability in llms via reinforcement learning.
\newblock \emph{arXiv preprint arXiv:2501.12948}, 2025.

\bibitem[Habermas(2015)]{habermas2015knowledge}
Habermas, J.
\newblock \emph{Knowledge and human interests}.
\newblock John Wiley \& Sons, 2015.

\bibitem[Hendrycks et~al.(2021)Hendrycks, Burns, Kadavath, Arora, et~al.]{hendrycks2021measuring}
Hendrycks, D., Burns, C., Kadavath, S., Arora, A., et~al.
\newblock Measuring mathematical problem solving with the math dataset.
\newblock In \emph{Thirty-fifth Conference on Neural Information Processing Systems Datasets and Benchmarks Track (Round 2)}, 2021.

\bibitem[Huang et~al.(2024)Huang, Chen, Mishra, Zheng, Yu, Song, and Zhou]{huanglarge}
Huang, J., Chen, X., Mishra, S., Zheng, H.~S., Yu, A.~W., Song, X., and Zhou, D.
\newblock Large language models cannot self-correct reasoning yet.
\newblock In \emph{The Twelfth International Conference on Learning Representations}, 2024.

\bibitem[Jonassen(2000)]{jonassen2000toward}
Jonassen, D.~H.
\newblock Toward a design theory of problem solving.
\newblock \emph{Educational technology research and development}, 48\penalty0 (4):\penalty0 63--85, 2000.

\bibitem[Koncel-Kedziorski et~al.(2016)Koncel-Kedziorski, Roy, Amini, Kushman, and Hajishirzi]{koncel2016mawps}
Koncel-Kedziorski, R., Roy, S., Amini, A., Kushman, N., and Hajishirzi, H.
\newblock Mawps: A math word problem repository.
\newblock In \emph{Proceedings of the 2016 conference of the north american chapter of the association for computational linguistics: human language technologies}, pp.\  1152--1157, 2016.

\bibitem[Leighton \& Gierl(2007)Leighton and Gierl]{leighton2007cognitive}
Leighton, J. and Gierl, M.
\newblock \emph{Cognitive diagnostic assessment for education: Theory and applications}.
\newblock Cambridge University Press, 2007.

\bibitem[Lesh \& Doerr(2003)Lesh and Doerr]{lesh2003foundations}
Lesh, R. and Doerr, H.~M.
\newblock Foundations of a models and modeling perspective on mathematics teaching, learning, and problem solving.
\newblock In \emph{Beyond constructivism}, pp.\  3--33. Routledge, 2003.

\bibitem[Li et~al.(2024{\natexlab{a}})Li, Hu, Huang, Zhuang, Liu, Zhu, Shi, and Lin]{li2024perteval}
Li, J., Hu, R., Huang, K., Zhuang, Y., Liu, Q., Zhu, M., Shi, X., and Lin, W.
\newblock Perteval: Unveiling real knowledge capacity of llms with knowledge-invariant perturbations.
\newblock In \emph{Advances in Neural Information Processing Systems}, 2024{\natexlab{a}}.

\bibitem[Li et~al.(2024{\natexlab{b}})Li, Cui, Zhao, Kong, and Bi]{li2024gsm}
Li, Q., Cui, L., Zhao, X., Kong, L., and Bi, W.
\newblock Gsm-plus: A comprehensive benchmark for evaluating the robustness of llms as mathematical problem solvers.
\newblock \emph{arXiv preprint arXiv:2402.19255}, 2024{\natexlab{b}}.

\bibitem[Liu et~al.(2024{\natexlab{a}})Liu, Feng, Wang, Wang, Liu, et~al.]{liu2024deepseek}
Liu, A., Feng, B., Wang, B., Wang, B., Liu, B., et~al.
\newblock Deepseek-v2: A strong, economical, and efficient mixture-of-experts language model.
\newblock \emph{arXiv preprint arXiv:2405.04434}, 2024{\natexlab{a}}.

\bibitem[Liu et~al.(2024{\natexlab{b}})Liu, Zheng, Qiao, Duan, et~al.]{liu2024mathbench}
Liu, H., Zheng, Z., Qiao, Y., Duan, H., et~al.
\newblock Mathbench: Evaluating the theory and application proficiency of llms with a hierarchical mathematics benchmark.
\newblock \emph{arXiv preprint arXiv:2405.12209}, 2024{\natexlab{b}}.

\bibitem[Liu et~al.(2023)Liu, Huang, Ma, Liu, Chen, Su, and Liu]{liu2023guiding}
Liu, J., Huang, Z., Ma, Z., Liu, Q., Chen, E., Su, T., and Liu, H.
\newblock Guiding mathematical reasoning via mastering commonsense formula knowledge.
\newblock In \emph{Proceedings of the 29th ACM SIGKDD Conference on Knowledge Discovery and Data Mining}, pp.\  1477--1488, 2023.

\bibitem[Liu et~al.(2025)Liu, Huang, Liu, Ma, Zhai, and Chen]{liu2025knowledge}
Liu, J., Huang, Z., Liu, Q., Ma, Z., Zhai, C., and Chen, E.
\newblock Knowledge-centered dual-process reasoning for math word problems with large language models.
\newblock \emph{IEEE Transactions on Knowledge and Data Engineering}, 2025.

\bibitem[Liu et~al.(2021)Liu, Huang, Yin, Chen, Xiong, Su, and Hu]{liu2021ekt}
Liu, Q., Huang, Z., Yin, Y., Chen, E., Xiong, H., Su, Y., and Hu, G.
\newblock Ekt: Exercise-aware knowledge tracing for student performance prediction.
\newblock \emph{IEEE Transactions on Knowledge and Data Engineering}, 33\penalty0 (1):\penalty0 100--115, 2021.

\bibitem[Lu et~al.(2024)Lu, Bansal, Xia, Liu, Li, et~al.]{lumathvista}
Lu, P., Bansal, H., Xia, T., Liu, J., Li, C., et~al.
\newblock Mathvista: Evaluating mathematical reasoning of foundation models in visual contexts.
\newblock In \emph{The Twelfth International Conference on Learning Representations}, 2024.

\bibitem[Ma et~al.(2025)Ma, Huang, Liu, Wang, Zhao, and Li]{ma2025automated}
Ma, Z., Huang, Z., Liu, J., Wang, M., Zhao, H., and Li, X.
\newblock Automated creation of reusable and diverse toolsets for enhancing llm reasoning.
\newblock In \emph{Proceedings of the AAAI Conference on Artificial Intelligence}, volume~39, pp.\  24821--24830, 2025.

\bibitem[Mao et~al.()Mao, Kim, and Zhou]{mao2024champ}
Mao, Y., Kim, Y., and Zhou, Y.
\newblock Champ: A competition-level dataset for fine-grained analyses of llms' mathematical reasoning capabilities.
\newblock In \emph{The 3rd Workshop on Mathematical Reasoning and AI at NeurIPS'23}.

\bibitem[Meta(2024)]{meta2024introducing}
Meta, A.
\newblock Introducing meta llama 3: The most capable openly available llm to date.
\newblock \emph{Meta AI}, 2024.

\bibitem[Min et~al.(2023)Min, Ross, Sulem, Veyseh, Nguyen, Sainz, Agirre, Heintz, and Roth]{min2023recent}
Min, B., Ross, H., Sulem, E., Veyseh, A. P.~B., Nguyen, T.~H., Sainz, O., Agirre, E., Heintz, I., and Roth, D.
\newblock Recent advances in natural language processing via large pre-trained language models: A survey.
\newblock \emph{ACM Computing Surveys}, 56\penalty0 (2):\penalty0 1--40, 2023.

\bibitem[MistralAITeam(2023)]{mistral2023mixtral}
MistralAITeam.
\newblock Mixtral-8x7b-v0.1.
\newblock \url{https://huggingface.co/mistralai/Mixtral-8x7B-v0.1}, 2023.

\bibitem[OpenAI(2023)]{chatgpt2023}
OpenAI.
\newblock \url{https://chatgpt.com/}, 2023.

\bibitem[Polya(2014)]{polya2014solve}
Polya, G.
\newblock How to solve it: A new aspect of mathematical method.
\newblock In \emph{How to solve it}. Princeton university press, 2014.

\bibitem[Rips(1994)]{rips1994psychology}
Rips, L.~J.
\newblock \emph{The psychology of proof: Deductive reasoning in human thinking}.
\newblock Mit Press, 1994.

\bibitem[Schoenfeld(2014)]{schoenfeld2014mathematical}
Schoenfeld, A.~H.
\newblock \emph{Mathematical problem solving}.
\newblock Elsevier, 2014.

\bibitem[Sun et~al.(2023)Sun, Li, Li, Wu, Guo, Zhang, and Wang]{sun2023text}
Sun, X., Li, X., Li, J., Wu, F., Guo, S., Zhang, T., and Wang, G.
\newblock Text classification via large language models.
\newblock In \emph{Findings of the Association for Computational Linguistics: EMNLP 2023}, pp.\  8990--9005, 2023.

\bibitem[Sweller(1988)]{sweller1988cognitive}
Sweller, J.
\newblock Cognitive load during problem solving: Effects on learning.
\newblock \emph{Cognitive science}, 12\penalty0 (2):\penalty0 257--285, 1988.

\bibitem[Team et~al.(2023)Team, Anil, Borgeaud, Wu, et~al.]{team2023gemini}
Team, G., Anil, R., Borgeaud, S., Wu, Y., et~al.
\newblock Gemini: a family of highly capable multimodal models.
\newblock \emph{arXiv preprint arXiv:2312.11805}, 2023.

\bibitem[Touvron et~al.(2023)Touvron, Martin, Stone, Albert, Almahairi, et~al.]{touvron2023llama}
Touvron, H., Martin, L., Stone, K., Albert, P., Almahairi, A., et~al.
\newblock Llama 2: Open foundation and fine-tuned chat models.
\newblock \emph{arXiv preprint arXiv:2307.09288}, 2023.

\bibitem[Wang et~al.(2023{\natexlab{a}})Wang, Lyu, Ji, Zhang, Yu, Shi, and Tu]{wang2023document}
Wang, L., Lyu, C., Ji, T., Zhang, Z., Yu, D., Shi, S., and Tu, Z.
\newblock Document-level machine translation with large language models.
\newblock In \emph{Proceedings of the 2023 Conference on Empirical Methods in Natural Language Processing}, pp.\  16646--16661, 2023{\natexlab{a}}.

\bibitem[Wang et~al.(2023{\natexlab{b}})Wang, Wei, Schuurmans, Le, et~al.]{wangself}
Wang, X., Wei, J., Schuurmans, D., Le, Q.~V., et~al.
\newblock Self-consistency improves chain of thought reasoning in language models.
\newblock In \emph{The Eleventh International Conference on Learning Representations}, 2023{\natexlab{b}}.

\bibitem[Wei et~al.(2022)Wei, Wang, Schuurmans, Bosma, Xia, Chi, Le, Zhou, et~al.]{wei2022chain}
Wei, J., Wang, X., Schuurmans, D., Bosma, M., Xia, F., Chi, E., Le, Q.~V., Zhou, D., et~al.
\newblock Chain-of-thought prompting elicits reasoning in large language models.
\newblock \emph{Advances in neural information processing systems}, 35:\penalty0 24824--24837, 2022.

\bibitem[Weng et~al.(2023)Weng, Zhu, Xia, Li, He, Liu, Sun, Liu, and Zhao]{weng2023large}
Weng, Y., Zhu, M., Xia, F., Li, B., He, S., Liu, S., Sun, B., Liu, K., and Zhao, J.
\newblock Large language models are better reasoners with self-verification.
\newblock In \emph{Findings of the Association for Computational Linguistics: EMNLP 2023}, pp.\  2550--2575, 2023.

\bibitem[Xu et~al.(2025)Xu, Xiao, Chao, Huang, Yang, and Wang]{xu2024can}
Xu, X., Xiao, T., Chao, Z., Huang, Z., Yang, C., and Wang, Y.
\newblock Can llms solve longer math word problems better?
\newblock In \emph{The Thirteenth International Conference on Learning Representations}, 2025.

\bibitem[Xue et~al.(2024)Xue, Huang, Liu, Lin, Ning, Jin, Li, and Liu]{xue2024decompose}
Xue, S., Huang, Z., Liu, J., Lin, X., Ning, Y., Jin, B., Li, X., and Liu, Q.
\newblock Decompose, analyze and rethink: Solving intricate problems with human-like reasoning cycle.
\newblock \emph{Advances in Neural Information Processing Systems}, 37:\penalty0 357--385, 2024.

\bibitem[Yao et~al.(2024)Yao, Yu, Zhao, Shafran, Griffiths, Cao, and Narasimhan]{yao2024tree}
Yao, S., Yu, D., Zhao, J., Shafran, I., Griffiths, T., Cao, Y., and Narasimhan, K.
\newblock Tree of thoughts: Deliberate problem solving with large language models.
\newblock \emph{Advances in Neural Information Processing Systems}, 36, 2024.

\bibitem[Yu et~al.(2024)Yu, Jiang, Shi, Jincheng, Liu, et~al.]{yumetamath}
Yu, L., Jiang, W., Shi, H., Jincheng, Y., Liu, Z., et~al.
\newblock Metamath: Bootstrap your own mathematical questions for large language models.
\newblock In \emph{The Twelfth International Conference on Learning Representations}, 2024.

\bibitem[Zhang et~al.(2024{\natexlab{a}})Zhang, Da, Lee, Robinson, Wu, Song, Zhao, Raja, Zhuang, Slack, et~al.]{zhang2024careful}
Zhang, H., Da, J., Lee, D., Robinson, V., Wu, C., Song, W., Zhao, T., Raja, P., Zhuang, C., Slack, D., et~al.
\newblock A careful examination of large language model performance on grade school arithmetic.
\newblock \emph{Advances in Neural Information Processing Systems}, 37:\penalty0 46819--46836, 2024{\natexlab{a}}.

\bibitem[Zhang et~al.(2024{\natexlab{b}})Zhang, Li, Zhang, Yin, Liu, and Moshfeghi]{zhang2024geoeval}
Zhang, J., Li, Z., Zhang, M., Yin, F., Liu, C., and Moshfeghi, Y.
\newblock Geoeval: benchmark for evaluating llms and multi-modal models on geometry problem-solving.
\newblock \emph{arXiv preprint arXiv:2402.10104}, 2024{\natexlab{b}}.

\bibitem[Zhang et~al.(2025{\natexlab{a}})Zhang, Wang, Ren, Jiang, Wang, and Liu]{zhang2025ratt}
Zhang, J., Wang, X., Ren, W., Jiang, L., Wang, D., and Liu, K.
\newblock Ratt: A thought structure for coherent and correct llm reasoning.
\newblock In \emph{Proceedings of the AAAI Conference on Artificial Intelligence}, volume~39, pp.\  26733--26741, 2025{\natexlab{a}}.

\bibitem[Zhang et~al.(2025{\natexlab{b}})Zhang, Zhang, Wu, and Zhao]{zhang2025gctn}
Zhang, L., Zhang, W., Wu, L., and Zhao, H.
\newblock Gctn: Graph competitive transfer network for cross-domain multi-behavior prediction.
\newblock \emph{IEEE Transactions on Knowledge and Data Engineering}, 2025{\natexlab{b}}.

\bibitem[Zhang et~al.(2024{\natexlab{c}})Zhang, Deng, Liu, Pan, and Bing]{zhang2024sentiment}
Zhang, W., Deng, Y., Liu, B., Pan, S., and Bing, L.
\newblock Sentiment analysis in the era of large language models: A reality check.
\newblock In \emph{Findings of the Association for Computational Linguistics: NAACL 2024}, pp.\  3881--3906, 2024{\natexlab{c}}.

\bibitem[Zhao et~al.(2024{\natexlab{a}})Zhao, Chen, Yang, Liu, Deng, Cai, Wang, Yin, and Du]{zhao2024explainability}
Zhao, H., Chen, H., Yang, F., Liu, N., Deng, H., Cai, H., Wang, S., Yin, D., and Du, M.
\newblock Explainability for large language models: A survey.
\newblock \emph{ACM Transactions on Intelligent Systems and Technology}, 15\penalty0 (2):\penalty0 1--38, 2024{\natexlab{a}}.

\bibitem[Zhao et~al.(2024{\natexlab{b}})Zhao, Wu, Shan, Jin, Sui, Liu, Feng, Li, and Zhang]{zhao2024comprehensive}
Zhao, H., Wu, L., Shan, Y., Jin, Z., Sui, Y., Liu, Z., Feng, N., Li, M., and Zhang, W.
\newblock A comprehensive survey of large language models in management: Applications, challenges, and opportunities.
\newblock \emph{Challenges, and Opportunities (August 14, 2024)}, 2024{\natexlab{b}}.

\bibitem[Zhao et~al.(2023)Zhao, Zhou, Li, Tang, Wang, Hou, Min, Zhang, Zhang, Dong, et~al.]{zhao2023survey}
Zhao, W.~X., Zhou, K., Li, J., Tang, T., Wang, X., Hou, Y., Min, Y., Zhang, B., Zhang, J., Dong, Z., et~al.
\newblock A survey of large language models.
\newblock \emph{arXiv preprint arXiv:2303.18223}, 2023.

\bibitem[Zhu et~al.(2024)Zhu, Wang, Zhao, Xu, and Xie]{zhu2024dynamic}
Zhu, K., Wang, J., Zhao, Q., Xu, R., and Xie, X.
\newblock Dynamic evaluation of large language models by meta probing agents.
\newblock In \emph{Forty-first International Conference on Machine Learning}, 2024.

\end{thebibliography}
\bibliographystyle{icml2025}

\newpage
\appendix
\begin{table*}[htbp]
    \centering
    \scriptsize
    \renewcommand{\arraystretch}{1.15} 
    \begin{tabular}{|m{1.5cm}|m{2.5cm}|m{5.1cm}|m{2.2cm}|}
        \hline
        \textbf{Stages} & \textbf{Dimensions} & \textbf{Example of Inquiry $q_i$} & \textbf{Pass} \\
        \hline
        \multirow{8}{1.5cm}{\raggedright Problem \\ Comprehension} 
        &  \textbf{Dimension 1:} \newline Sentence Paraphrasing & 
        \underline{Jacob} had $\$21$. \underline{Emily shared} half of her $\$100$ \underline{with him}. How much money \underline{does Jacob} have now? & Answer Correctly \\
        \cline{2-4}
        &  \textbf{Dimension 2:} \newline Sentence Disruption & 
        $\$21$ Ali had half of $\$100$ him Leila her gave now? does Ali much How have & Identify ``Unsolvable'' \\
        \cline{2-4}
        &  \textbf{Dimension 3:} \newline Missing Condition & 
        Ali had \underline{some money}. Leila gave him half of her \underline{money}. How much does Ali have now? & Identify ``Unsolvable'' \\
        \cline{2-4}
        &  \textbf{Dimension 4:} \newline Redundant Condition & 
        Ali had $\$21$. Leila gave him half of her $\$100$. \underline{Before meeting with Leila, Ali had already counted} \underline{his money twice to make sure it was correct.} How much does Ali have now? & Answer Correctly \\
        \hline
        
        \multirow{6}{1.5cm}{\raggedright Problem \\ Solving}
        &  \textbf{Dimension 5:} \newline Analogical Reasoning & 
        Tom had $\$21$ comic books. Jerry traded him half of his collection of $\$100$ comic books. How many comic books does Tom have now? & Answer Correctly \\
        \cline{2-4}
        &  \textbf{Dimension 6:} \newline Numerical Transformation & 
        Ali had \underline{$\$30$}. Leila gave him half of her \underline{$\$120$}. How much does Ali have now? & Answer Correctly \\
        \cline{2-4}
        &  \textbf{Dimension 7:} \newline Knowledge Redefinition & 
        \underline{Assume ``half'' means one-third of the given amount}, solve the following problem: Ali had $\$21$. Leila gave him half of her $\$100$. How much does Ali have now? & Answer Correctly \\
        \hline
        
        \multirow{4}{1.5cm}{\raggedright Solution \\ Summarization} 
        &  \textbf{Dimension 8:} \newline Intermediate Step Questioning & 
        Given the mathematical problem: Ali had $\$21$. Leila gave him half of her $\$100$. How much does Ali have now? please answer my following question: \underline{Why does Ali now have $\$71$?} & Answer Correctly \\
        \cline{2-4}
        &  \textbf{Dimension 9:} \newline Backward Reasoning & 
        In the problem, ``Ali had $\$21$. Leila gave him half of her \underline{$\alpha$}, where $\alpha$ is an unknown total amount of money Leila had. How much does Ali have now?'', if Ali now has $\$71$, \underline{what is the value of $\alpha$?} & Answer Correctly \\
        \hline
    \end{tabular}
    \caption{The 3 cognitive stages and 9 dimensions in our CogMath. ``Pass'' refers to the type of LLM response that is considered to pass the inquiry $q_i$ of the given dimension.}\label{cogmath_summarize}
\end{table*}
\section{An Example of CogMath}\label{cogmath_details}
We present in Table~\ref{cogmath_summarize} the inquiries across nine dimensions in CogMath for mathematical problem ``Ali had \$21. Leila gave him half of her \$100. How much does Ali have now?''.

\section{Prompts in CogMath Framework}\label{append_prompts}
The prompts for all agents across the 9 dimensions are presented in Figures~\ref{append_prompts}.1.1 to~\ref{append_prompts}.8.2. Notably, in CogMath, the expected answers for Dimensions 1 to 4 are the original answers $A$ of problem $P$, so we omit the corresponding \emph{Reference} agents for these dimensions. For Dimension 2, the \emph{Inquiry} agent automatically disrupts the word order in each clause according to rules, and this process does not require a special prompt or a \emph{Judge} agent for evaluation. Hence, all agents for Dimension 2 are omitted here. As for Dimension 9, as shown in Figure A.8.1, its \emph{Inquiry} agent also automatically determines the answer for inquiry $q_9$ (marked with ``[]''), so there is no need to design an additional \emph{Reference} agent prompt, which is therefore omitted.

\section{Implementation Details}\label{implementation_details}
All the \emph{Inquiry} agents, \emph{Reference} agents, and \emph{Judge} agents are implemented with GPT-4. Besides, the maximum number of iterations for \emph{Inquiry} agent is set to $\delta=10$. If after 10 iterations, we still fail to obtain a satisfactory inquiry, we consider the problem to be unsuitable to be evaluated from that dimension. For such problems, we omit consideration of that dimension during the evaluation.

We apply CogMath on two of the most representative mathematical benchmarks, GSM8K~\citep{cobbe2021training} and MATH~\citep{hendrycks2021measuring}, along with our constructed MExam dataset. GSM8K is an elementary-level math word problem dataset that primarily involves basic understanding and reasoning. MATH is a high school competition-level dataset, consisting of 7 subcategories, such as algebra and geometry. MExam is composed of 6,353 questions manually collected from real exams in 50 Chinese exercise books, which covers the full K-12 mathematics curriculum. For GSM8K and MATH, since their training sets may have already been used in the training process of current LLMs, we apply CogMath on their public test sets, which contain 1,319 and 5,000 questions, respectively.
\section{Evaluation of DeepSeek-R1}
Due to the cost and rate limitations of API calls, here we evaluate the current SOTA DeepSeek-R1~\cite{guo2025deepseek} on the most widely used GSM8K and MATH datasets in Table~\ref{exp_r1}. First, compared to Table~\ref{performance_all}, DeepSeek-R1 achieves the best performance among all evaluated LLMs, both in ``Vanilla'' and CogMath framework. This shows its superior mathematical reasoning capabilities. Second, based on the performance gap (marked as $\Delta$), DeepSeek-R1 still exhibits a certain degree of overestimation, highlighting the necessity of our proposed evaluation from the human cognitive perspective. Third, similar to other advanced LLMs, DeepSeek-R1 encounters the most challenges in Stage 2 (i.e., \emph{Problem Solving}). Further analysis reveals that its main weakness lies in Dimension 7 (Knowledge Redefinition), with a \emph{Relative Pass Rate} (RPR) of 0.617. This supports the conclusion that current LLMs rely on fixed memorization rather than adapting knowledge flexibly. Fourth, compared to DeepSeek-V2.5, DeepSeek-R1 improves significantly in Stage 3 (\emph{Solution Summarization}), suggesting a deeper understanding of reasoning process.
\begin{table*}[t]
\small
\centering
\setlength{\tabcolsep}{5.5pt} 
\begin{tabular}{c|c|cccccccc|c}
\toprule[1.5pt]
\multicolumn{2}{l|}{}    & \multicolumn{8}{c|}{\footnotesize \textbf{MATH}} & \multirow{2}{*}{\centering \footnotesize \textbf{GSM8K}} \\
\cline{3-10} \multicolumn{2}{l|}{} & \footnotesize \textbf{Avg} & \footnotesize \textbf{Alg} & \footnotesize \textbf{Count} & \footnotesize \textbf{Geo} & \footnotesize \textbf{Itmd} & \footnotesize \textbf{Num} & \footnotesize \textbf{Pre-Alg} & \footnotesize \textbf{Pre-Cal} & \\ 
\hline
\multirow{6}{*}{\scriptsize DeepSeek-R1} & \scriptsize Vanilla & 0.982 & 0.992 & 0.994 & 0.956 & 0.979 & 0.980 & 0.985 & 0.972 & 0.967  \\ 
& \scriptsize CogMath                          & 0.448 & 0.581 & 0.443 & 0.307 & 0.295 & 0.413 & 0.604 & 0.326 & 0.703  \\ 
& $\Delta$                                  &  \cellcolor[gray]{0.9}-0.534 & \cellcolor[gray]{0.9}-0.411 & \cellcolor[gray]{0.9}-0.551 & \cellcolor[gray]{0.9}-0.649 & \cellcolor[gray]{0.9}-0.684 & \cellcolor[gray]{0.9}-0.567 & \cellcolor[gray]{0.9}-0.381 & \cellcolor[gray]{0.9}-0.646 & \cellcolor[gray]{0.9}-0.264 \\ 
\cline{2-11} & \scriptsize \emph{Stage} 1 & 0.863 & 0.942 & 0.831 & 0.737 & 0.837 & 0.881 & 0.875 & 0.837 & 0.897 \\ 
&\scriptsize \emph{Stage} 2                        & \textbf{0.575} & \textbf{0.715} & \textbf{0.557} & \textbf{0.441} & \textbf{0.405} & \textbf{0.544} & \textbf{0.738} & \textbf{0.456} &0.848 \\ 
&\scriptsize \emph{Stage} 3                        & 0.753 & 0.808 & 0.773 & 0.637 & 0.694 & 0.743 & 0.815 & 0.725 & \textbf{0.856} \\ 
\bottomrule[1.5pt]
\end{tabular}
\caption{Performance of DeepSeek-R1.}\label{exp_r1}
\end{table*}
\section{Dimension-level Effects of LLM enhancement methods}\label{dimension_llm}
Here we present the effects of ICL on each dimension. As defined in Section~\ref{exp_prompt}, the higher \emph{Pass Rate} represents better performance.
\begin{table*}[t]
    \small
    \renewcommand{\arraystretch}{1.1}
    \centering
    \begin{tabular}{c|c|c|c|c|c|c|c|c|c|c}
    \toprule[1.5pt]
    \multicolumn{2}{c|}{MATH} &  D.1 & D.2 & D.3 &  D.4 & D.5 & D.6 &  D.7 & D.8 & D.9 \\
    \hline
    \multirow{3}{*}{GPT-4} & CogMath & 0.737 & 0.589 & 0.596 & 0.732 & 0.695 & 0.632 & 0.611 & 0.943 & 0.727 \\
                         & CogMath(ICL) & 0.722 & 0.557 & 0.645 & 0.714 & 0.638 & 0.681 & 0.625 & 0.930 & 0.720 \\
                         & $\Delta$     &-1.5\%	& -3.2\% & -4.9\% &-1.8\% &-5.7\% &+4.9\% &+1.4\% &-1.3\% &-0.7\% \\
   \hline
    \multirow{3}{*}{GPT-3.5} & CogMath & 0.486 & 0.666 & 0.700 & 0.486 & 0.509 & 0.457 & 0.379 & 0.823 & 0.556 \\
                         & CogMath(ICL) & 0.504 & 0.582 & 0.733 & 0.498 & 0.494 & 0.499 & 0.434 & 0.832 & 0.559 \\
                         & $\Delta$     &+1.8\%	& -8.4\% & +3.3\% &+1.2\% &-1.5\% &+4.2\% &+5.5\% &+0.9\% &+0.3\% \\
   \hline
    \multirow{3}{*}{Gemini-1.5} & CogMath & 0.602 & 0.569 & 0.764 & 0.591 & 0.603 & 0.531 & 0.467 & 0.887 & 0.696 \\
                         & CogMath(ICL) & 0.595 & 0.662 & 0.730 & 0.589 & 0.543 & 0.545 & 0.466 & 0.870 & 0.619 \\
                         & $\Delta$     &-0.7\%	& +9.3\%& -3.4\%& -0.2\%& -6.0\%& +1.4\%& -0.1\%& -1.7\%& -7.7\% \\
   \bottomrule[1.5pt]
   \multicolumn{2}{c|}{GSM8K} &  D.1 & D.2 & D.3 &  D.4 & D.5 & D.6 &  D.7 & D.8 & D.9 \\
    \hline
    \multirow{3}{*}{GPT-4} & CogMath & 0.886 & 0.692 & 0.657 & 0.946 & 0.930 & 0.921 & 0.792 & 0.976 & 0.828 \\
                         & CogMath(ICL) & 0.889 & 0.662 & 0.754 & 0.943 & 0.920 & 0.928 & 0.801 & 0.972 & 0.853 \\
                         & $\Delta$     &+0.3\% & -3.0\% & +9.7\% & -0.3\% & -1.0\% & +0.7\% & +0.9\% & -0.4\% & +2.5\% \\
   \hline
    \multirow{3}{*}{GPT-3.5} & CogMath & 0.730 & 0.741 & 0.728 & 0.816 & 0.833 & 0.792 & 0.589 & 0.899 & 0.668 \\
                         & CogMath(ICL) & 0.778 & 0.640 & 0.773 & 0.802 & 0.810 & 0.826 & 0.592 & 0.901 & 0.704 \\
                         & $\Delta$     &+4.8\%	& -10.1\% & +4.5\%& -1.4\%&-2.3\%& +3.4\%& +0.3\%& +0.2\%& +3.6\% \\
   \hline
    \multirow{3}{*}{Gemini-1.5} & CogMath & 0.773 & 0.730 & 0.985 & 0.821 & 0.890 & 0.873 & 0.672 & 0.907 & 0.786 \\
                         & CogMath(ICL) & 0.807 & 0.763 & 0.859 & 0.895 & 0.873 & 0.868 & 0.697 & 0.861 & 0.773 \\
                         & $\Delta$     & +3.4\% & +3.3\% & -12.6\% & +7.4\% & -1.7\% & -0.5\% & +2.5\% & -4.6\% & -1.3\% \\
   \bottomrule[1.5pt]
    \end{tabular}
    \caption{Dimension-level Performances of In-Context Learning.}
    \label{perform_prompt_dimension}
\end{table*}

From Table~\ref{perform_prompt_dimension}, we observe that Dimensions 6 and 7 show consistently stable improvements with the introduction of ICL for all LLMs. This suggests that ICL has a significant positive impact on the reasoning abilities in handling numerical transformations and knowledge redefinition. Second, Dimension 5 benefits the least from ICL, and it consistently experiences negative effects. This may be due to the reasoning process in the demonstration diverging from that of the original problem, which could disrupt the model's performance in analogical reasoning (i.e., reasoning that follows the same process as the original problem). Third, for other dimensions, the effect of ICL varies across different models. For instance, in Dimension 3, GPT-4 and GPT-3.5 show improvements, while Gemini-1.5 shows a decline. This highlights the differing robustness and capabilities of various models, providing insights into potential weaknesses and future development directions of different LLMs.
\section{Discussion}\label{append_discuss}
First, our CogMath framework is highly generalizable, as it does not rely on specific problem types or formats, making it applicable to testing LLMs' cognitive abilities in other mathematical tasks, such as theorem proving. Besides, beyond evaluating individually, we can also compose multiple dimensions for assessing human-like multi-behaviors~\cite{zhang2025gctn} and deeper cognitive diagnosis~\cite{liu2021ekt}. Second, our framework can be easily extended to tasks in other domains. For instance, in visual reasoning tasks, a \emph{visual comprehension} stage could be added into our framework, along with dimensions like image perturbation to evaluate the capabilities and robustness of visual LLMs like GPT-4v. Third, through experiments in Sections~\ref{exp_main} to~\ref{exp_error}, we have conducted a detailed examination of LLMs' mastery across different dimensions, providing valuable insights for future model improvements. For example, as observed in Section~\ref{section_cog_dimension}, existing LLMs may exhibit an ``over-correction'' behavior when faced with unsolvable problems. To address this, we need to introduce critical thinking mechanisms that enable them to reconsider the fundamental nature of each problem~\cite{zhang2025ratt}, rather than merely imitating patterns from training data. Lastly, from the results of Section~\ref{exp_prompt}, we found that CoT and ICL may not fundamentally improve the mathematical capabilities of LLMs. However, these techniques have been shown to enhance performance in many NLP tasks. Therefore, we believe that understanding the underlying mechanisms of these methods from a theoretical perspective remains a critical research question.
\begin{figure}[t]
\centering
\vspace{-55pt}
\setlength{\abovecaptionskip}{2pt}
\includegraphics[width=0.99\linewidth]{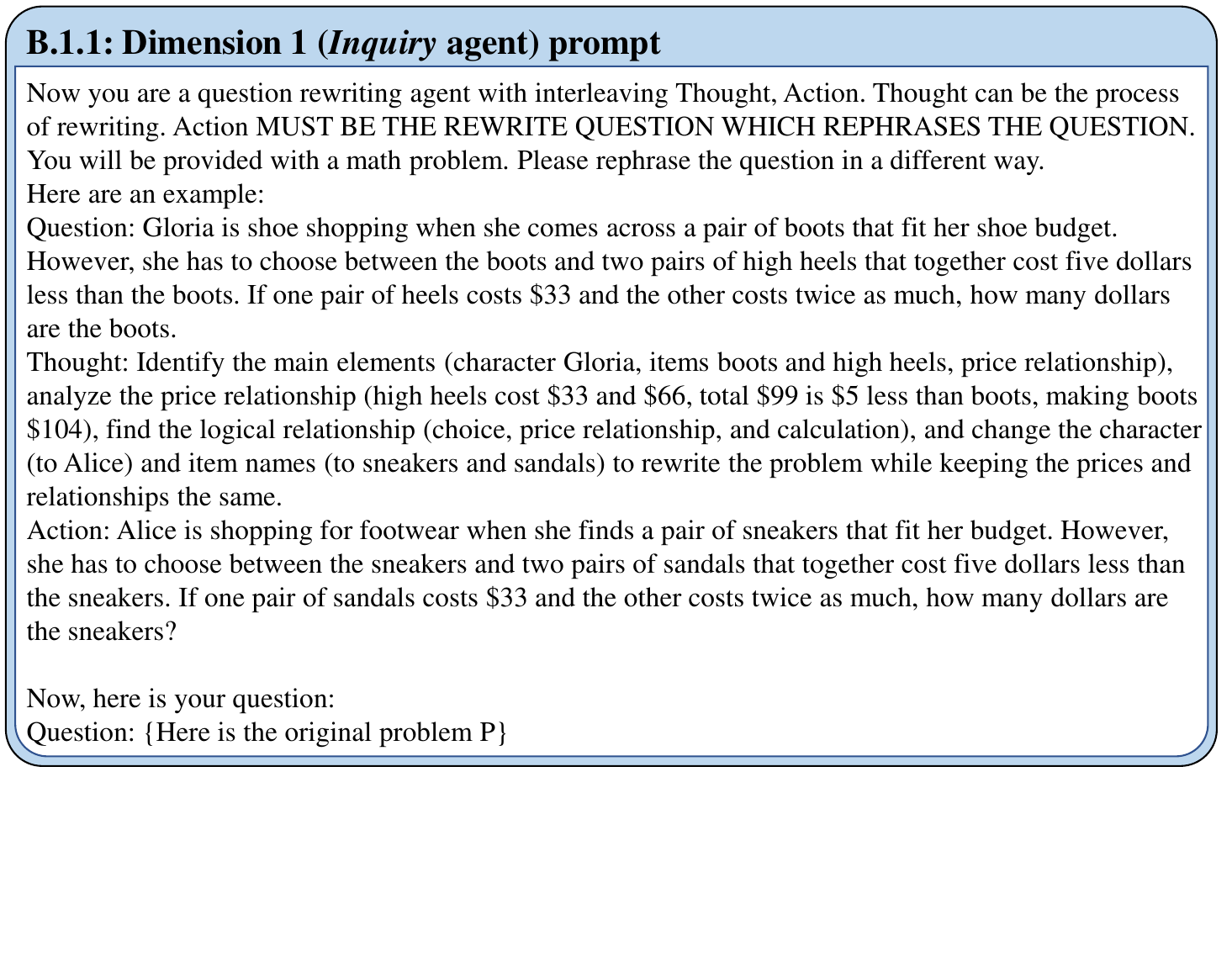}
\end{figure}

\begin{figure}[h]
\centering
\setlength{\abovecaptionskip}{2pt}
\includegraphics[width=0.99\linewidth]{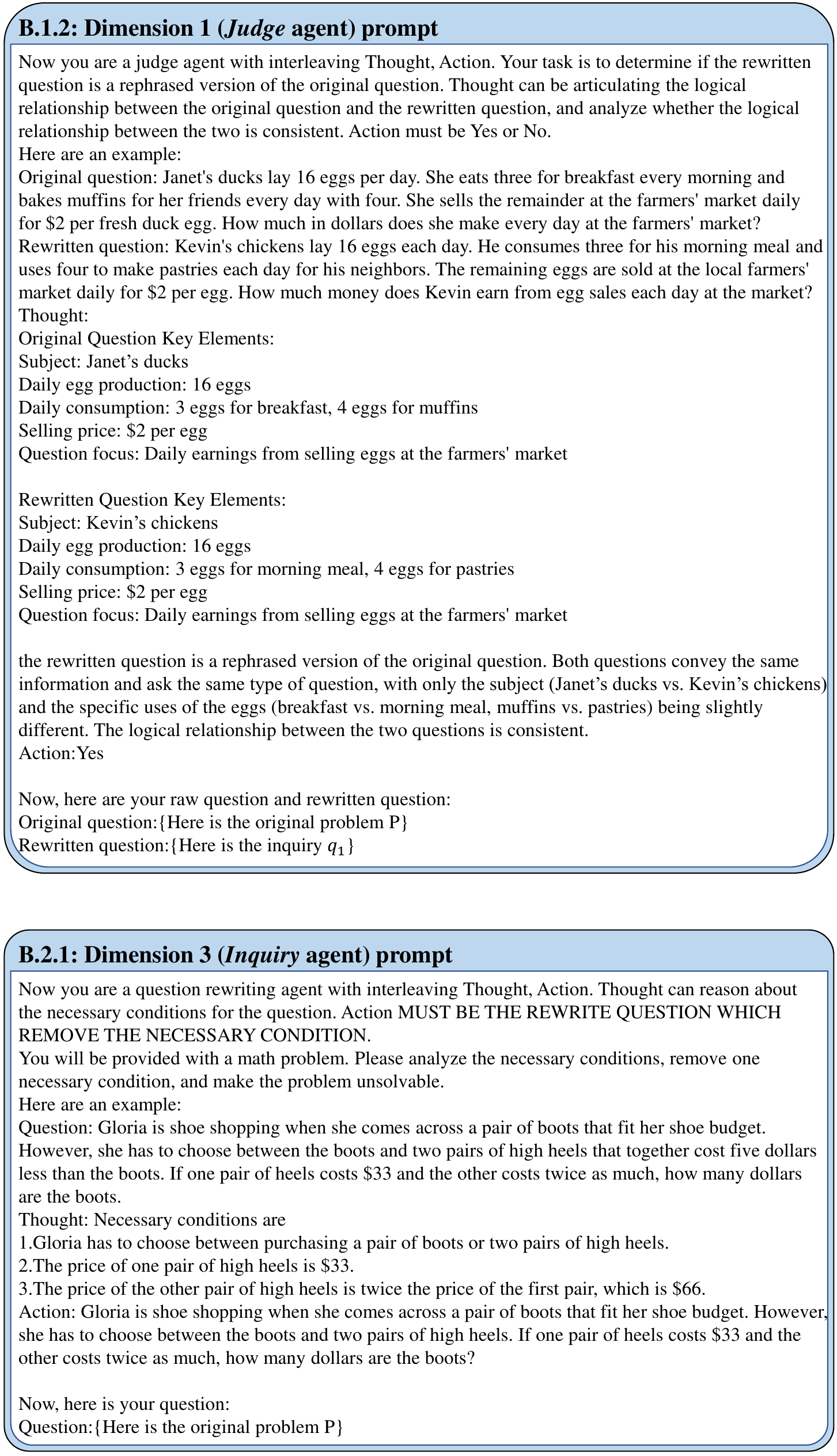}
\vspace{-12pt}
\end{figure}

\begin{figure}[!ht]
\centering
\setlength{\abovecaptionskip}{2pt}
\includegraphics[width=0.99\linewidth]{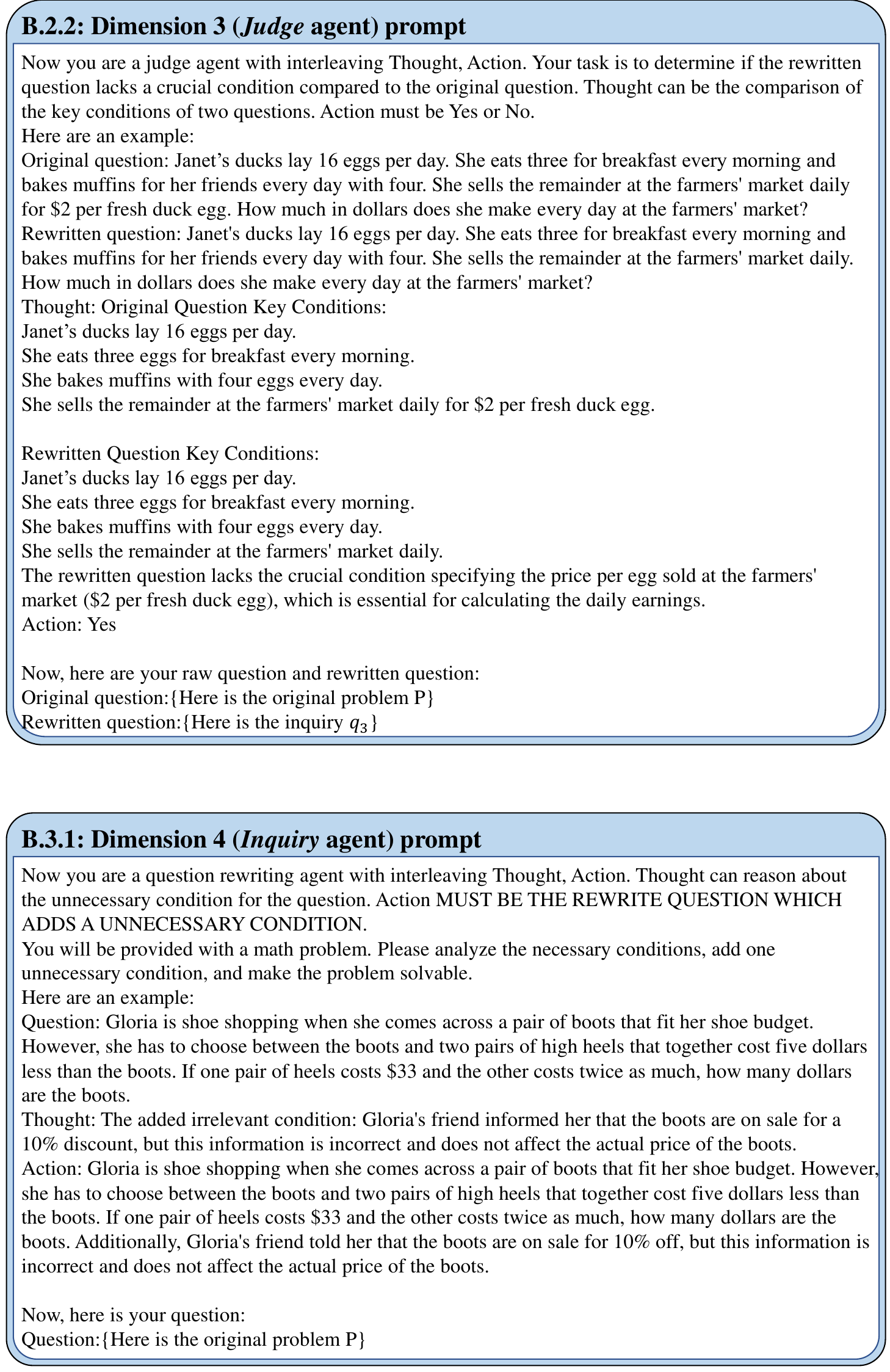}
\end{figure}

\begin{figure}[t]
\centering
\setlength{\abovecaptionskip}{2pt}
\includegraphics[width=0.99\linewidth]{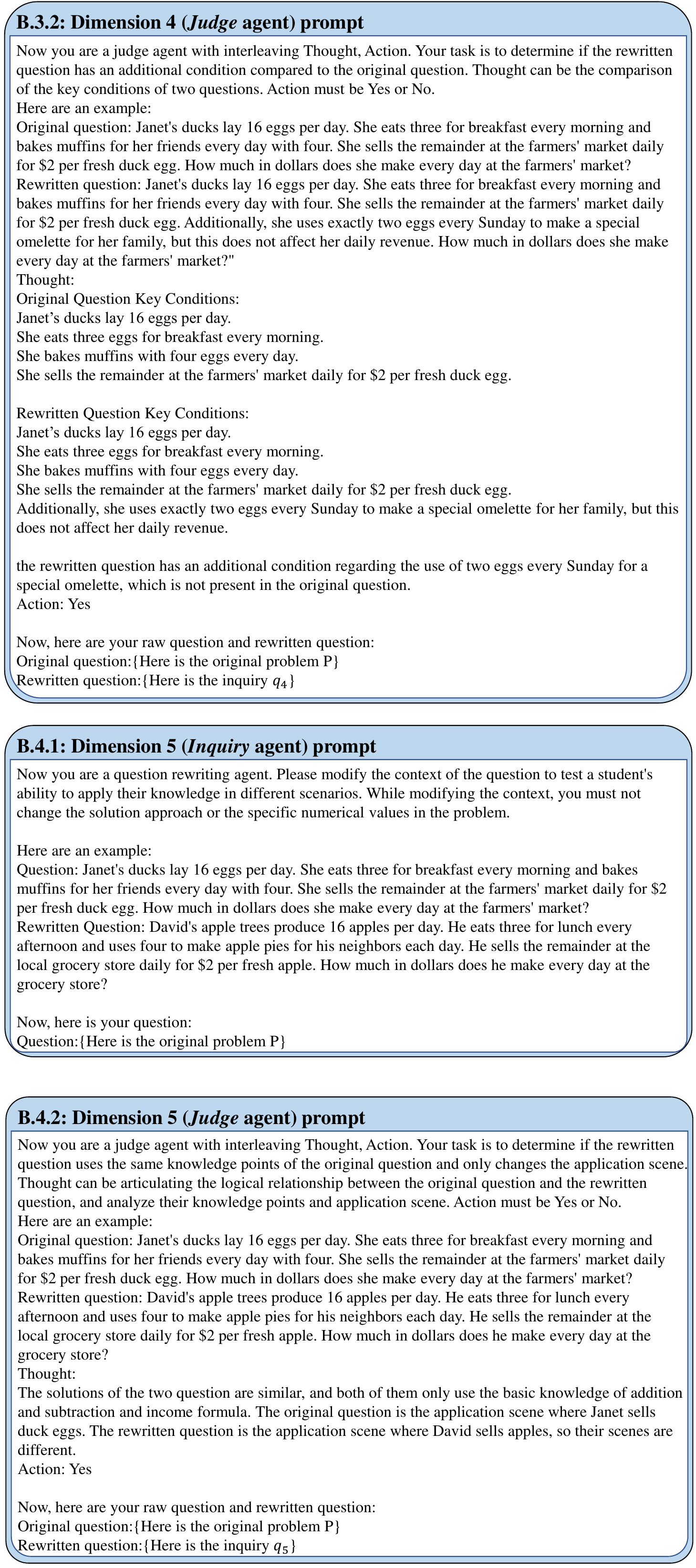}
\end{figure}

\begin{figure}[t]
\centering
\setlength{\abovecaptionskip}{2pt}
\includegraphics[width=0.99\linewidth]{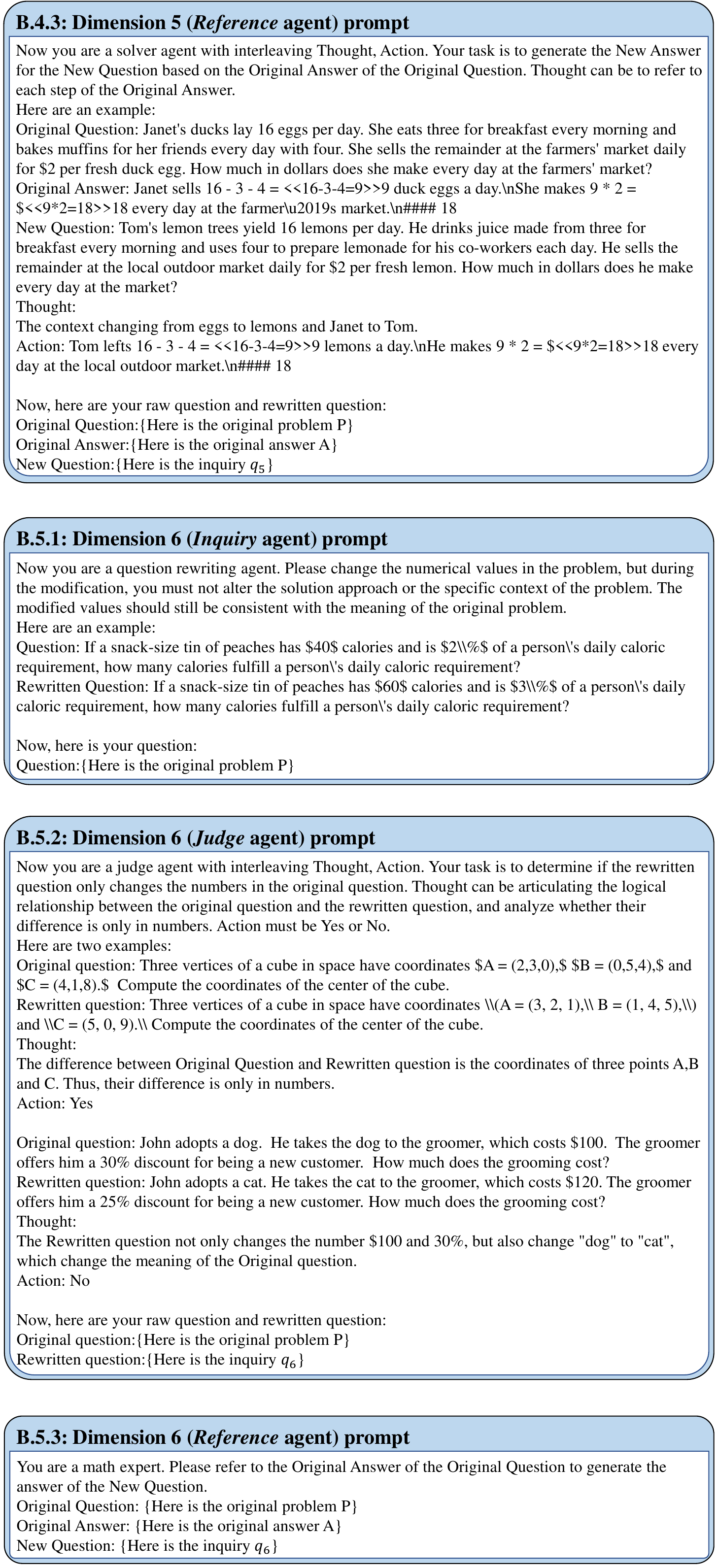}
\end{figure}

\begin{figure}[!ht]
\centering
\setlength{\abovecaptionskip}{2pt}
\includegraphics[width=0.99\linewidth]{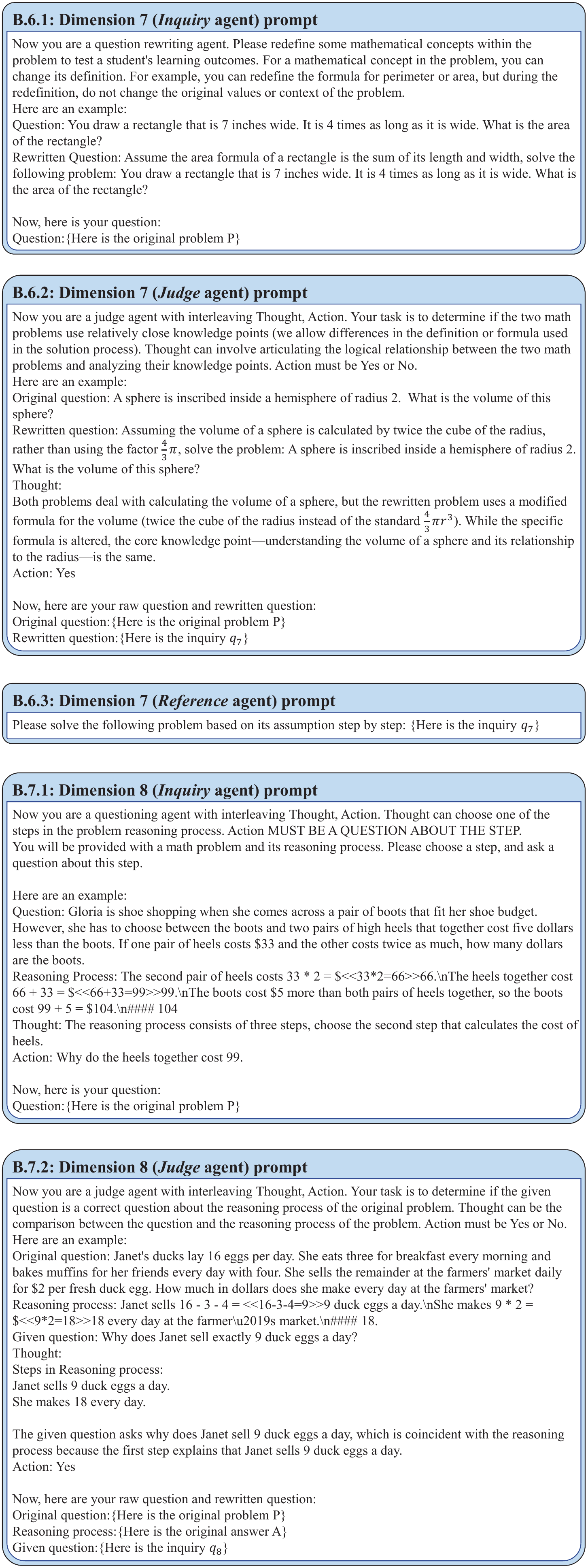}
\end{figure}

\begin{figure}[t]
\centering
\setlength{\abovecaptionskip}{2pt}
\includegraphics[width=0.99\linewidth]{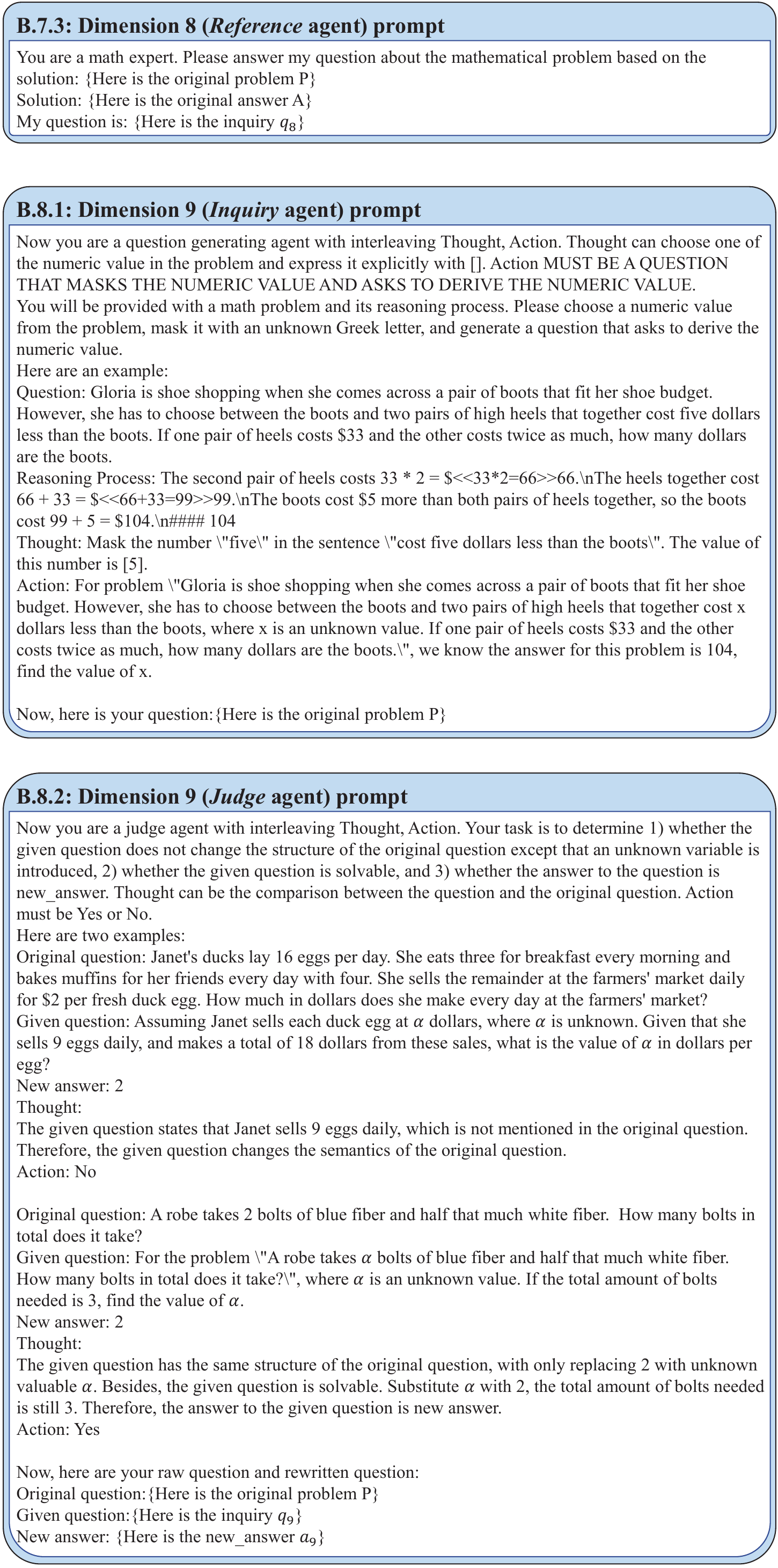}
\end{figure}

\end{document}